\pdfoutput=1
\documentclass[11pt]{article}
\usepackage[dvipsnames, svgnames, x11names]{xcolor}
\usepackage{acl2023}

\usepackage{times}
\usepackage{latexsym}

\usepackage[T1]{fontenc}
\usepackage[utf8]{inputenc}

\usepackage{microtype}

\usepackage{booktabs}
\usepackage{multirow}
\usepackage{amsmath}
\usepackage{amssymb}
\usepackage{xcolor}
\usepackage{pifont}
\usepackage{bbm}

\usepackage{tikz}
\usepackage{subfigure}
\usepackage{pgfplots}
\usepackage{makecell,rotating,diagbox}
\usepackage{verbatim}
\usepackage{colortbl}

\usepackage{algorithm}
\usepackage{algorithmic}
\usepackage{graphicx}
\usepackage{arydshln}
\usepackage{wasysym}

\usepackage[normalem]{ulem}

\usepackage{mathrsfs}

\usetikzlibrary{plotmarks}
\usetikzlibrary{patterns}
\usetikzlibrary{pgfplots.groupplots}

\usetikzlibrary{plotmarks}
\usetikzlibrary{arrows}
\usetikzlibrary{decorations}
\usetikzlibrary{decorations.pathreplacing}
\usetikzlibrary{decorations.markings}
\usetikzlibrary{backgrounds}
\usetikzlibrary{fit}
\usetikzlibrary{positioning}
\usetikzlibrary{calc}
\usetikzlibrary{patterns}
\usetikzlibrary{shadows}
\usetikzlibrary[shapes.multipart]
\usetikzlibrary{pgfplots.groupplots}
\usetikzlibrary{arrows.meta}

\newdimen\base
\newdimen\baseh
\newdimen\basew
\baseh=\base
\basew=\base

\newdimen\legendmargin % left/right margin
\newdimen\legendwidth % symbol width
\newdimen\legendsep % item-item gap

\definecolor{byte1}{HTML}{3156A4}
\definecolor{byte2}{HTML}{487FC0}
\definecolor{byte3}{HTML}{2CB7C1}
\definecolor{byte4}{HTML}{81CAC7}

\definecolor{ugreen}{cmyk}{1,0,1,0.498}
\definecolor{lyyblue}{cmyk}{0.8278,0.3333,0,0.2941}
\definecolor{lyygreen}{cmyk}{0.6813,0,0.725,0.3725}
\definecolor{lyyred}{cmyk}{0,0.8855,0.8767,0.1098}
\definecolor{dblue}{cmyk}{1,0.5487,0,0.5569}
\definecolor{lypurple}{HTML}{e0c2c0}
\definecolor{lygreen}{HTML}{eff67b}
\definecolor{lyblue}{HTML}{d5ddef}
\definecolor{lyblue1}{HTML}{bfe2fe}
\definecolor{lyyellow}{HTML}{fdfab5}
\definecolor{lyyellow1}{HTML}{E1DBE5}
\definecolor{lyypink}{HTML}{ffe0db}
\definecolor{lypink}{HTML}{fecdcb}
\definecolor{lyred}{HTML}{b71a3b}
\definecolor{lygrey}{HTML}{dedfe4}
\definecolor{lyorange}{HTML}{FFEBCD}
\definecolor{zshgreen}{rgb}{0.5607,0.6902,0.5725}
\definecolor{zshorange}{HTML}{fa6d1d}
\definecolor{zshblue}{rgb}{0.6,0.6,1}
\definecolor{zshpink}{rgb}{1,0.6,0.6}

\definecolor{back1}{HTML}{F2F2F2}
\definecolor{back2}{HTML}{FEF9D6}

\newcommand{\tab}[1]{Table \ref{#1}}%
\newcommand{\fig}[1]{Fig. \ref{#1}}%
\newcommand{\softmax}{$\mathrm{SoftMax}$}
\newcommand{\relu}{$\mathrm{ReLU}$}

\usepackage{mathtools}

\title{MobileNMT: One System that Translates in 15MB and 30ms}
\title{MobileNMT: Enabling Translation in 15MB and 30ms}
\author{
  Ye Lin\textsuperscript{1}\thanks{\ \ This work is done during the internship at ByteDance.},
  Xiaohui Wang\textsuperscript{2},
  Zhexi Zhang\textsuperscript{2},
  Mingxuan Wang\textsuperscript{2},
  Tong Xiao\textsuperscript{1,3}\thanks{\ \ Corresponding author.},
  Jingbo Zhu\textsuperscript{1,3} \\
  \textsuperscript{1}NLP Lab, School of Computer Science and Engineering, \\
    Northeastern University, Shenyang, China \\
  \textsuperscript{2} ByteDance \\
  \textsuperscript{3}NiuTrans Research, Shenyang, China \\
  {\tt \{linye2015\}@outlook.com}\\
  {\tt \{wangxiaohui.neo,zhangzhexi,wangmingxuan.89\}@bytedance.com}\\
  {\tt \{xiaotong,zhujingbo\}@mail.neu.edu.cn} \\
}

\begin{document}
\maketitle

\begin{abstract}

Deploying NMT models on mobile devices is essential for privacy, low latency, and offline scenarios. For high model capacity, NMT models are rather large. Running these models on devices is challenging with limited storage, memory, computation, and power consumption. Existing work either only focuses on a single metric such as FLOPs or general engine which is not good at auto-regressive decoding. In this paper, we present MobileNMT, a system that can translate in 15MB and 30ms on devices. We propose a series of principles for model compression when combined with quantization. Further, we implement an engine that is friendly to INT8 and decoding. With the co-design of model and engine, compared with the existing system, we speed up 47.0$\times$ and save 99.5\% of memory with only 11.6\% loss of BLEU.
The code is publicly available
at https://github.com/zjersey/Lightseq-ARM.
% \textcolor{red}{Our code will be publicly available after the anonymity period.}      
\end{abstract}

\section{Introduction}

% With the rapid development of modern processing units and distributed training, it is easy to train large models on devices such as GPUs.
% However, constrained by limited hardware resources, these large-scale models can be difficult to deploy on handheld devices such as smartphones.

As a classic subfield of natural language processing, neural machine translation (NMT) has achieved great success in recent years.
Most of the studies focus on improving the accuracy of large machine translation systems, ignoring whether such models are easy to be deployed in real-world scenarios.
% It can produce results with surprising accuracy and fluency.
% To ensure the accuracy of the NMT model, the number of parameters is usually set to 100$\sim$300M, which leads to a large model that may require more than 1GB of disk space when storing in 32-bit floating point format.
% At the same time, deploying such models requires significant computational and memory resources.
% For example, to translate a sentence with 30 words, \textcolor{red}{Transformer-base/big?}

Here we adopt four metrics to evaluate whether an NMT model is deployment-friendly. 
(1) \textbf{Model size} is the most important metric in model compression \cite{DBLP:journals/corr/HanMD15}.
(2) \textbf{Floating-point operations (FLOPs)} is commonly used to evaluate computational complexity in neural architecture design.
(3) \textbf{Memory} or \textbf{Memory mapped I/O (MMI/O)} reflects the memory requirements of the real running system.
(4) \textbf{Decoding speed} depends on many realistic factors such as engine implementation and the power of avaliable processors.

% \begin{table}[!t]
%   \centering
%   \small
%   \renewcommand\arraystretch{1.1}
%   \setlength{\tabcolsep}{0.4mm}{
%     \begin{tabular}{l|c|c|c|c|c}
%       \hline
%       \multicolumn{1}{c|}{\multirow{2}{*}{System}} &
%       \multicolumn{1}{c|}{\multirow{1}{*}{Size}} &
%       {\multirow{1}{*}{FLOPs}} &
%       \multicolumn{1}{c|}{\multirow{1}{*}{MMI/O}} &
%       \multicolumn{1}{c|}{Speed} &
%       \multicolumn{1}{c}{\multirow{2}{*}{BLEU}} \\
%       % \cline{3-4}
%       & (MB) & (G) & (M) &   
%       (sent./s) \\
%       \hline
%       \multirow{1}{*}{Transformer-big} & 872 & 6.4 & 210 & - & 28.36 \\
%       \multirow{1}{*}{Transformer-base} & 260 & 1.9 & 66 & - & 27.40 \\
%       \multirow{1}{*}{MobileNMT-20MB} & 20 & 0.6 & 22 & - & 27.09 \\
%       \multirow{1}{*}{MobileNMT-10MB} & 10 & 0.3 & 11 & - & 25.08 \\
%       \hline
%     \end{tabular}
%   \caption{The comparison of different models (FLOPs and MMI/O are measured on a sample with src/tgt length of 30 and 40K/8K vocabulary.).}
%   \label{tab:comparison}
%   }
% \end{table}

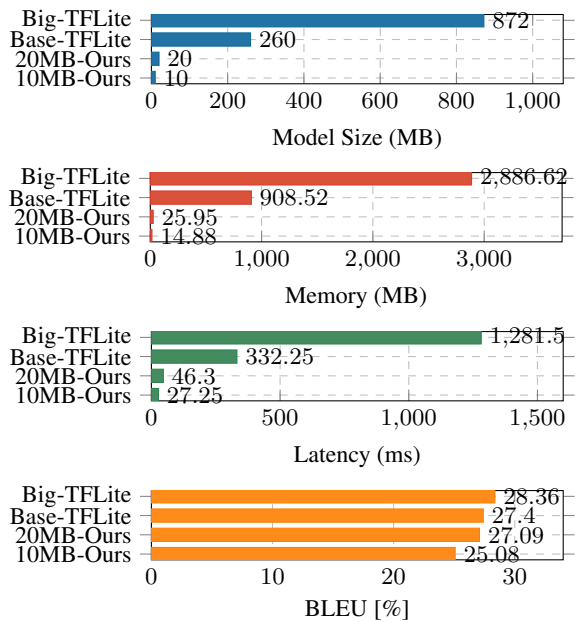
\begin{figure}[t]
  \centering
  \small
  \begin{tikzpicture}
    \begin{axis}[
      width=7.0cm, height=2.5cm, 
      xbar, 
      grid style=dashed,
      ymajorgrids=true,
      xmajorgrids=true,
      ylabel near ticks,
      xlabel near ticks,
      xlabel={Model Size (MB)},
      xmin=0,
      xmax=1080,
      bar width=5pt,
      symbolic y coords={10MB-Ours,20MB-Ours,Base-TFLite,Big-TFLite},
      ytick=data,
      nodes near coords,
      nodes near coords style={font=\small}]
      \addplot[draw=lyyblue,fill=lyyblue!90] coordinates {
      (10,10MB-Ours)(20,20MB-Ours)(260,Base-TFLite)(872,Big-TFLite)};
    \end{axis}
  \end{tikzpicture}
  \begin{tikzpicture}
    \begin{axis}[
      width=7.0cm, height=2.5cm, 
      xbar, 
      grid style=dashed,
      ymajorgrids=true,
      xmajorgrids=true,
      ylabel near ticks,
      xlabel near ticks,
      xlabel={Memory (MB)},
      xmin=0,
      xmax=3700,
      bar width=5pt,
      symbolic y coords={10MB-Ours,20MB-Ours,Base-TFLite,Big-TFLite},
      ytick=data,
      nodes near coords,
      nodes near coords style={font=\small}]
      \addplot[draw=lyyred,fill=lyyred!90] coordinates {
      (14.88,10MB-Ours)(25.95,20MB-Ours)(908.52,Base-TFLite)(2886.62,Big-TFLite)};
    \end{axis}
  \end{tikzpicture}
  \begin{tikzpicture}
    \begin{axis}[
      width=7.0cm, height=2.5cm, 
      xbar, 
      grid style=dashed,
      ymajorgrids=true,
      xmajorgrids=true,
      ylabel near ticks,
      xlabel near ticks,
      xlabel={Latency (ms)},
      xmin=0,
      xmax=1600,
      bar width=5pt,
      symbolic y coords={10MB-Ours,20MB-Ours,Base-TFLite,Big-TFLite},
      ytick=data,
      nodes near coords,
      nodes near coords style={font=\small}]
      \addplot[draw=lyygreen,fill=lyygreen!90] coordinates {
      (27.25,10MB-Ours)(46.30,20MB-Ours)(332.25,Base-TFLite)(1281.50,Big-TFLite)};
    \end{axis}
  \end{tikzpicture}
  \begin{tikzpicture}
    \begin{axis}[
      width=7.0cm, height=2.5cm, 
      xbar, 
      grid style=dashed,
      ymajorgrids=true,
      xmajorgrids=true,
      ylabel near ticks,
      xlabel near ticks,
      xlabel={BLEU [\%]},
      xmin=0,
      xmax=34,
      bar width=5pt,
      symbolic y coords={10MB-Ours,20MB-Ours,Base-TFLite,Big-TFLite},
      ytick=data,
      nodes near coords,
      nodes near coords style={font=\small}]
      \addplot[draw=orange,fill=orange!90] coordinates {
      (25.08,10MB-Ours)(27.09,20MB-Ours)(27.40,Base-TFLite)(28.36,Big-TFLite)};
    \end{axis}
  \end{tikzpicture}
  \caption{These metrics are measured on Google Pixel 4. Each result is the average of 200 runs on a sample of src/tgt length 30.}
  \label{fig:comparison}
\end{figure}

\begin{figure*}[t!]
  \centering
  \small
  % \hspace*{\fill}
  \begin{tikzpicture}
    \begin{axis}[
      width=0.75\columnwidth,height=0.42\columnwidth,
      yticklabel style={/pgf/number format/fixed,/pgf/number format/precision=1},
      ylabel={BLEU [\%]},
      ylabel near ticks,
      xlabel={\# Params of Base (M)},
      xlabel near ticks,
      enlargelimits=0.1,
      % symbolic x coords={45,50,55,60,65},
      xmajorgrids=true,
      ymajorgrids=true,
      grid style=dashed,
      xtick=data,
      every tick label/.append style={font=\small},
      legend columns=1,
      legend pos=south east,
      legend style={font=\small},
    ]
      \addplot [lyyred,thick,mark=*] coordinates {
        (64,27.40) (60,27.04) (55,26.91) (49,26.33) (47,25.46)
      };\addlegendentry{Scaling E}
      \addplot [lyyblue,thick,mark=square*] coordinates {
        (64,27.40) (59,27.44) (54,27.09) (49,26.35) (47,26.17)
      };\addlegendentry{Scaling V}
    \end{axis}
  \end{tikzpicture}
  % \hfill
  \begin{tikzpicture}
    \begin{axis}[
      width=0.75\columnwidth,height=0.42\columnwidth,
      yticklabel style={/pgf/number format/fixed,/pgf/number format/precision=1},
      ylabel={BLEU [\%]},
      ylabel near ticks,
      xlabel={\# Params of Small (M)},
      xlabel near ticks,
      enlargelimits=0.1,
      % symbolic x coords={45,50,55,60,65},
      xmajorgrids=true,
      ymajorgrids=true,
      grid style=dashed,
      xtick=data,
      every tick label/.append style={font=\small},
      legend columns=1,
      legend pos=south east,
      legend style={font=\small},
    ]
      \addplot [lyyred,thick,mark=*] coordinates {
        (21,24.20) (19,23.93) (16,23.76) (14,22.99) (12,21.03)
      };\addlegendentry{Scaling E}
      \addplot [lyyblue,thick,mark=square*] coordinates {
        (21,24.20) (19,24.40) (16,24.00) (14,22.97) (12,22.49)
      };\addlegendentry{Scaling V}
    \end{axis}
  \end{tikzpicture}
  \begin{tikzpicture}
    \begin{axis}[
      width=0.75\columnwidth,height=0.42\columnwidth,
      yticklabel style={/pgf/number format/fixed,/pgf/number format/precision=1},
      ylabel={BLEU [\%]},
      ylabel near ticks,
      xlabel={\# Params of Tiny (M)},
      xlabel near ticks,
      enlargelimits=0.1,
      % symbolic x coords={45,50,55,60,65},
      xmajorgrids=true,
      ymajorgrids=true,
      grid style=dashed,
      xtick=data,
      every tick label/.append style={font=\small},
      legend columns=1,
      legend pos=south east,
      legend style={font=\small},
    ]
      \addplot [lyyred,thick,mark=*] coordinates {
        (8,20.97) (7,20.91) (5,20.31) (4,18.53) (3,14.48)
      };\addlegendentry{Scaling E}
      \addplot [lyyblue,thick,mark=square*] coordinates {
        (8,20.97) (7,20.80) (5,19.97) (4,18.69) (3,17.10)
      };\addlegendentry{Scaling V}
    \end{axis}
  \end{tikzpicture}
  \begin{tikzpicture}
    \begin{axis}[
      width=0.74\columnwidth,height=0.42\columnwidth,
      yticklabel style={/pgf/number format/fixed,/pgf/number format/precision=1},
      ylabel={BLEU [\%]},
      ylabel near ticks,
      xlabel={\# Params of Base (M)},
      xlabel near ticks,
      enlargelimits=0.1,
      % symbolic x coords={45,50,55,60,65},
      xmajorgrids=true,
      ymajorgrids=true,
      grid style=dashed,
      xtick=data,
      every tick label/.append style={font=\small},
      legend columns=1,
      legend pos=south east,
      legend style={font=\small},
    ]
      \addplot [lyyred,thick,mark=*] coordinates {
        (28.2,25.41) (35.5,25.90) (42,26.20) (64.5,27.40)
      };\addlegendentry{Sharing}
      \addplot [lyyblue,thick,mark=square*] coordinates {
        (28.2,25.18) (35.7,25.96) (43.7,26.13) (64.5,27.40) 
      };\addlegendentry{Width}
    \end{axis}
  \end{tikzpicture}
  % \hfill
  \begin{tikzpicture}
    \begin{axis}[
      width=0.74\columnwidth,height=0.42\columnwidth,
      yticklabel style={/pgf/number format/fixed,/pgf/number format/precision=1},
      ylabel={BLEU [\%]},
      ylabel near ticks,
      xlabel={\# Params of Small (M)},
      xlabel near ticks,
      enlargelimits=0.1,
      % symbolic x coords={45,50,55,60,65},
      xmajorgrids=true,
      ymajorgrids=true,
      grid style=dashed,
      xtick=data,
      every tick label/.append style={font=\small},
      legend columns=1,
      legend pos=south east,
      legend style={font=\small},
    ]
      \addplot [lyyred,thick,mark=*] coordinates {
        (12.2,22.37) (14.1,23.37) (15.9,23.87) (21.5,24.20)
      };\addlegendentry{Sharing}
      \addplot [lyyblue,thick,mark=square*] coordinates {
        (12.8,22.62) (14.0,22.94) (15.2,23.34) (21.5,24.20)
      };\addlegendentry{Width}
    \end{axis}
  \end{tikzpicture}
  \begin{tikzpicture}
    \begin{axis}[
      width=0.74\columnwidth,height=0.42\columnwidth,
      yticklabel style={/pgf/number format/fixed,/pgf/number format/precision=1},
      ylabel={BLEU [\%]},
      ylabel near ticks,
      xlabel={\# Params of Tiny (M)},
      xlabel near ticks,
      enlargelimits=0.1,
      % symbolic x coords={45,50,55,60,65},
      xmajorgrids=true,
      ymajorgrids=true,
      grid style=dashed,
      xtick=data,
      every tick label/.append style={font=\small},
      legend columns=1,
      legend pos=south east,
      legend style={font=\small},
    ]
      \addplot [lyyred,thick,mark=*] coordinates {
        (5.7,17.75) (6.1,19.94) (6.6,20.00) (8.0,20.97) 
      };\addlegendentry{Sharing}
      \addplot [lyyblue,thick,mark=square*] coordinates {
        (5.6,18.60) (5.8,19.14) (6.3,19.79) (8.0,20.97) 
      };\addlegendentry{Width}
    \end{axis}
  \end{tikzpicture}
  % \hspace*{\fill}
  \caption{Model performance of different methods in Section \ref{sec:architecture} and Section \ref{sec:training}
  (Scaling E: scaling embedding dimension; Scaling V: scaling vocabulary size; Sharing: cross-layer parameter sharing; Width: reducing model width).
  \uwave{Scaling V performs better than Scaling E. Width performs nearly the same with Sharing.}}
  \label{fig:embedding_encdec}
\end{figure*}
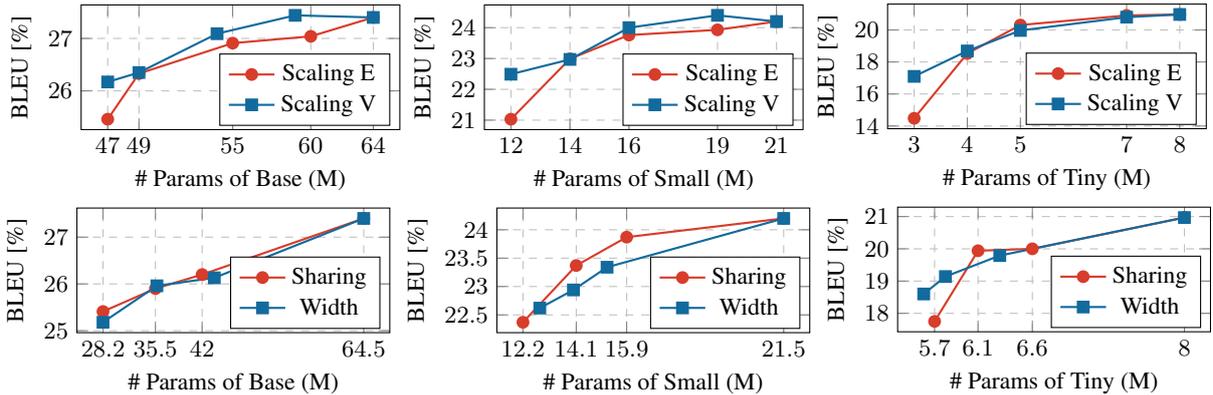

In this paper, we propose MobileNMT, a Transformer-based machine translation system that can translate in 15MB and 30ms.
First, we propose three principles for designing parameter-limited MT models: 
1) To compress embedding, reducing vocabulary size is simple and effective compared to embedding factorization;
2) To compress the encoder and decoder, reducing the model width is much more efficient in computation and memory than cross-layer parameter sharing;
3) Encoder depth is very important to ensure accuracy.
To achieve higher accuracy, we adjust the training hyperparameters according to the newly designed structure, and adopt sequence-level knowledge distillation.
For industrial deployment, we optimize general matrix multiplication (GEMM) and memory in our own inference engine and use the 8-bit integer for storage and computation.
As shown in \tab{fig:comparison}, the 10MB MobileNMT achieves 88.4\% performance of Transformer-big with only 1.1\% size and runs 47.0$\times$ faster on decoding, which can be easily deployed and used.

Our contributions are summarized as follows:
\begin{itemize}
  \item We propose three principles for parameter-limited MT models to make more efficient use of computation and memory resources.
  \item We adjust training strategies according to the newly designed structure to achieve higher translation accuracy.
  \item We develop a mobile inference engine to bridge the gap between industrial practice and theoretical research.
\end{itemize}

 % \section{Parameters Assigning}
\section{Architecture Design Principles}
\label{sec:architecture}

For model compression and acceleration, most studies focus on a single metric such as model size or FLOPs, without considering the real-world applications. 
In this section, we consider four metrics including model size, FLOPs, memory usage, and decoding speed, and then propose three design principles for parameter-limited MT models.
We choose Transformer (Appendix \ref{sec:transformer}) as our baseline because of its great success in machine translation.
% We will introduce how to design the training strategies in Section \ref{sec:training}, and introduce the implementation of our engine in Section \ref{sec:engine}.

% \subsection{Parameters of the Embedding}
\subsection{Embedding Compression}

The vocabulary size $V$ usually reaches tens of thousands in NMT models \cite{DBLP:conf/wmt/AkhbardehABBCCC21}. 
The parameters can reach tens of millions and greatly affect the overall parameter efficiency. 
% Reducing the parameters of this part is important for designing parameter-efficient NMT models.

\textbf{Embedding Factorization (Scaling E).}
For model compression, embedding factorization has been widely studied \cite{DBLP:conf/iclr/LanCGGSS20,DBLP:conf/icml/GraveJCGJ17,DBLP:conf/iclr/BaevskiA19}.
% The embedding dimension $E$ and hidden dimension $H$ in the original Transformer model are set to be the same. 
To decouple the embedding dimension $E$ and hidden dimension $H$, it additionally introduces a trainable transformation weight $W^T\in\mathbb{R}^{E\times H}$, where $E\leq H$.
After factorization, the embedding parameters will be decreased from $O(V\times H)$ to $O(V\times E+E\times H)$.
% Suppose $E=H/2$, then the parameters can be compressed nearly twice.
% which can bring a very considerable parameter reduction.

\begin{table}[t!]
  \centering
  \small
  \renewcommand\arraystretch{1.0}
  \setlength{\tabcolsep}{0.0mm}
  \begin{tabular}{c|c|cccc|cccc|cccc}
      \hline
      \multicolumn{1}{c|}{\multirow{1}{*}{Module}} & Dim & \multicolumn{4}{c|}{Base} & \multicolumn{4}{c|}{Small} & \multicolumn{4}{c}{Tiny} \\
      \hline
      \multicolumn{1}{c|}{\multirow{3}{*}{Embed}} & Vocab & \multicolumn{1}{r}{\multirow{3}{*}{\Huge{$\left[\right.$}}} & 40,000 & \multirow{3}{*}{{\Huge$\left.\right]$}} & & \multicolumn{1}{r}{\multirow{3}{*}{\Huge{$\left[\right.$}}} & 40,000 & \multirow{3}{*}{{\Huge$\left.\right]$}} & & \multicolumn{1}{r}{\multirow{3}{*}{\Huge{$\left[\right.$}}} & 40,000 & \multirow{3}{*}{{\Huge$\left.\right]$}} \\
      \multicolumn{1}{c|}{\multirow{1}{*}{}} & Embed & & N/A & & $\times1$ & & N/A & & $\times1$ & & N/A & & $\times1$ \\
      \multicolumn{1}{c|}{\multirow{1}{*}{}} & Hidden & & 512 & & & & 256 & & & & 128 & & \\
      \hline
      \multicolumn{1}{c|}{\multirow{3}{*}{Encoder}} & Hidden & \multicolumn{1}{r}{\multirow{3}{*}{\Huge{$\left[\right.$}}} & 512 & \multirow{3}{*}{{\Huge$\left.\right]$}} & & \multicolumn{1}{r}{\multirow{3}{*}{\Huge{$\left[\right.$}}} & 256 & \multirow{3}{*}{{\Huge$\left.\right]$}} & & \multicolumn{1}{r}{\multirow{3}{*}{\Huge{$\left[\right.$}}} & 128 & \multirow{3}{*}{{\Huge$\left.\right]$}} \\
      \multicolumn{1}{c|}{\multirow{1}{*}{}} & Head & & 8 & & $\times6$ & & 4 & & $\times6$ & & 2 & & $\times6$ \\
      \multicolumn{1}{c|}{\multirow{1}{*}{}} & FFN & & 2048 & & & & 1024 & & & & 512 & & \\
      \hline
      \multicolumn{1}{c|}{\multirow{3}{*}{Decoder}} & Hidden & \multicolumn{1}{r}{\multirow{3}{*}{\Huge{$\left[\right.$}}} & 512 & \multirow{3}{*}{{\Huge$\left.\right]$}} & & \multicolumn{1}{r}{\multirow{3}{*}{\Huge{$\left[\right.$}}} & 256 & \multirow{3}{*}{{\Huge$\left.\right]$}} & & \multicolumn{1}{r}{\multirow{3}{*}{\Huge{$\left[\right.$}}} & 128 & \multirow{3}{*}{{\Huge$\left.\right]$}} \\
      \multicolumn{1}{c|}{\multirow{1}{*}{}} & Head & & 8 & & $\times6$ & & 4 & & $\times6$ & & 2 & & $\times6$ \\
      \multicolumn{1}{c|}{\multirow{1}{*}{}} & FFN & & 2048 & & & & 1024 & & & & 512 & & \\
      \hline
      \multicolumn{2}{c|}{\multirow{1}{*}{Params}} & \multicolumn{4}{c|}{\multirow{1}{*}{64.5M}} & \multicolumn{4}{c|}{\multirow{1}{*}{21.5M}} & \multicolumn{4}{c}{\multirow{1}{*}{8.0M}} \\
      \hline
  \end{tabular}
  \caption{The detailed settings of Base, Small and Tiny.}
  \label{tab:detailed}
\end{table}

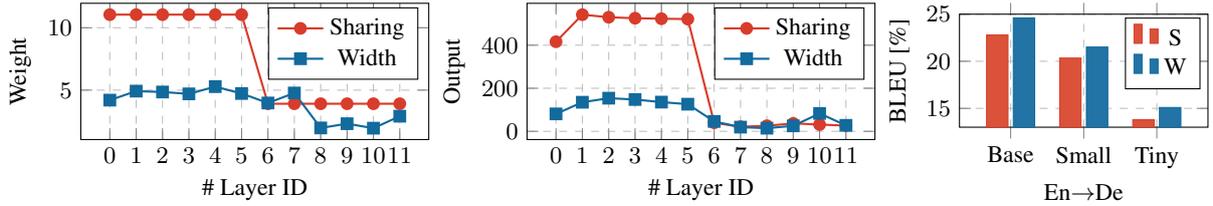
\begin{figure*}[t!]
  \centering
  \small
  % \hspace*{\fill}
  \begin{tikzpicture}
    \begin{axis}[
      width=0.8\columnwidth,height=0.44\columnwidth,
      yticklabel style={/pgf/number format/fixed,/pgf/number format/precision=1},
      ylabel={Weight},
      ylabel near ticks,
      xlabel={\# Layer ID},
      xlabel near ticks,
      enlargelimits=0.1,
      % symbolic x coords={45,50,55,60,65},
      xmajorgrids=true,
      ymajorgrids=true,
      grid style=dashed,
      xtick=data,
      every tick label/.append style={font=\small},
      legend columns=1,
      legend pos=north east,
      legend style={font=\small},
    ]
      \addplot [lyyred,thick,mark=*] coordinates {
        (0,11.05) (1,11.05) (2,11.05) (3,11.05) (4,11.05) (5,11.05) (6,3.91) (7,3.91) (8,3.91) (9,3.91) (10,3.91) (11,3.91)
      };\addlegendentry{Sharing}
      \addplot [lyyblue,thick,mark=square*] coordinates {
        (0,4.20) (1,4.92) (2,4.85) (3,4.70) (4,5.28) (5,4.73) (6,3.97) (7,4.77) (8,1.97) (9,2.31) (10,1.95) (11,2.91)
      };\addlegendentry{Width}
    \end{axis}
  \end{tikzpicture}
  % \hfill
  \begin{tikzpicture}
    \begin{axis}[
      width=0.8\columnwidth,height=0.44\columnwidth,
      yticklabel style={/pgf/number format/fixed,/pgf/number format/precision=1},
      ylabel={Output},
      ylabel near ticks,
      xlabel={\# Layer ID},
      xlabel near ticks,
      enlargelimits=0.1,
      % symbolic x coords={45,50,55,60,65},
      xmajorgrids=true,
      ymajorgrids=true,
      grid style=dashed,
      xtick=data,
      every tick label/.append style={font=\small},
      legend columns=1,
      legend pos=north east,
      % legendwidth=0.001\linewidth,
      % legendmargin=0.01\linewidth,
      % legendsep=0.008\linewidth,
      legend style={font=\small},
    ]
      \addplot [lyyred,thick,mark=*] coordinates {
        (0,416.09) (1,541.8) (2,529.8) (3,525.3) (4,522.9) (5,521.4) (6,39.675) (7,21.16) (8,25.85) (9,36.78) (10,31.40) (11,26.39)
      };\addlegendentry{Sharing}
      \addplot [lyyblue,thick,mark=square*] coordinates {
        (0,80.92) (1,135.0) (2,153.9) (3,147.7) (4,135.67) (5,126.0) (6,46.01) (7,19.34) (8,14.5) (9,25.39) (10,83.0) (11,27.65)
      };\addlegendentry{Width}
    \end{axis}
  \end{tikzpicture}
  \begin{tikzpicture}
    \begin{axis}[
      width=0.63\columnwidth,height=0.4\columnwidth,
      ybar, ymin=13,ymax=25,
      grid style=dashed,
      ymajorgrids=true,
      xmajorgrids=true,
      ylabel={BLEU [\%]},
      xlabel=En$\rightarrow$De,
      ylabel near ticks,
      xlabel near ticks,
      enlarge x limits=0.35,
      bar width=8pt,
      symbolic x coords={Base,Small,Tiny},
      xtick=data,
      nodes near coords style={font=\small},
      legend columns=1,
      legend pos=north east,
      legend style={font=\small},
      ]
      \addplot[draw=lyyred,fill=lyyred!90] coordinates {
      (Base,22.79)(Small,20.34)(Tiny,13.79)};\addlegendentry{S}
      \addplot[draw=lyyblue,fill=lyyblue!90] coordinates {
      (Base,24.60)(Small,21.51)(Tiny,15.09)};\addlegendentry{W}
    \end{axis}
  \end{tikzpicture}
  % \hspace*{\fill}
  \caption{The left two figures show weight and output ranges for each layer. 
  The right figure shows the model performance of Post Training Quantization (PTQ) in cross-layer parameter sharing vs. reducing model width. 
  \uwave{These figures show that reducing model width is more quantization-friendly than cross-layer parameter sharing.}}
  \label{fig:encdec_quantization}
\end{figure*}

\textbf{Reducing Vocabulary Size (Scaling V).}
A more direct way to compress embedding is to reduce the vocabulary size $V$.
To reduce the risk of out-of-vocabulary words, here we adopt Byte-Pair Encoding (BPE) \cite{DBLP:conf/acl/SennrichHB16a,DBLP:conf/wmt/OttEGA18,DBLP:conf/mtsummit/DingRD19,DBLP:journals/corr/abs-2008-07772}.
% BPE breaks words into subword units, it starts from the alphabet and merges characters into the most frequent subword units, then segments words in sentences by these merged subword units.
For most studies on machine translation, the adopted BPE merge operations range from 30$\sim$40K \cite{DBLP:conf/mtsummit/DingRD19}.
Volt proves that we can find a well-performing vocabulary with higher BLEU and smaller BPE merge operations \cite{DBLP:conf/acl/XuZGZL20}.
Experiments in \citet{DBLP:conf/emnlp/LinLXZ21}'work also show that smaller vocabularies may be better.

% \textbf{Reducing vocabulary size performs better.}
\textbf{Reducing Vocabulary Size Performs Better.}
To compare the two embedding compression methods, here we select three baseline models of different sizes. 
The model settings are shown in \tab{tab:detailed}.
% For further parameter-efficiency, here we share all embeddings in one matrix.
As shown in \tab{tab:comparison1}, the parameters and FLOPs are almost the same in these two methods.
As shown in the first row of \fig{fig:embedding_encdec}, compared to reducing vocabulary size, the model with embedding factorization performs poorly in most cases, especially when the parameters are limited. 

\begin{table}[!t]
  \centering
  \small
  \renewcommand\arraystretch{1.05}
  \setlength{\tabcolsep}{0.8mm}{
    \begin{tabular}{l|c|c|c|c|c|c|c}
      \hline
      \multicolumn{1}{c|}{\multirow{2}{*}{Metric}} &
      \multicolumn{3}{c|}{\multirow{1}{*}{Scaling E}} &
      \multicolumn{1}{c|}{\multirow{6}{*}{vs.}} &
      \multicolumn{3}{c}{\multirow{1}{*}{Scaling V}} \\
      \cline{2-4}
      \cline{6-8}
      & Base & Small & Tiny & & Base & Small & Tiny \\
      \cline{1-4}
      \cline{6-8}
      \multirow{1}{*}{Params (M)} & 47 & 12 & 3 & & 47 & 12 & 3 \\
      \multirow{1}{*}{FLOPs (G)} & 1.41 & 0.38 & 0.11 & & 1.41 & 0.38 & 0.11 \\
      \multirow{1}{*}{MMI/O (M)} & 48 & 15 & 6 & & 47 & 14 & 5 \\
      \multirow{1}{*}{BLEU} & 25.46 & 21.03 & 14.48 & & 26.17 & 22.49 & 17.10 \\
      \hline
      \hline
      \multicolumn{1}{c|}{\multirow{2}{*}{Metric}} &
      \multicolumn{3}{c|}{\multirow{1}{*}{Sharing}} &
      \multicolumn{1}{c|}{\multirow{6}{*}{vs.}} &
      \multicolumn{3}{c}{\multirow{1}{*}{Width}} \\
      \cline{2-4}
      \cline{6-8}
      & Base & Small & Tiny & & Base & Small & Tiny \\
      \cline{1-4}
      \cline{6-8}
      \multirow{1}{*}{Params (M)} & 28 & 12 & 6 & & 28 & 12 & 6 \\
      \multirow{1}{*}{FLOPs (G)} & 1.95 & 0.65 & 0.24 & & 0.85 & 0.38 & 0.17 \\
      \multirow{1}{*}{MMI/O (M)} & 66 & 24 & 10 & & 30 & 15 & 7 \\
      \multirow{1}{*}{BLEU} & 25.41 & 22.37 & 17.75 & & 25.18 & 22.62 & 18.60 \\
      \hline
    \end{tabular}
  \caption{Parameters, FLOPs, and model performance (FLOPs and MMI/O are estimated on a sample with src/tgt length of 30.). \uwave{For embedding compression, reducing vocabulary size (Scaling V) is more simple and effective.
  For encoder/decoder compression, reducing model width (Width) is more efficient in computation and memory.}}
  \label{tab:comparison1}
  }
\end{table}

\subsection{Encoder/Decoder Compression}
% \subsection{Reducing Model Width Instead of Cross-Layer Parameter Sharing}

For encoder and decoder compression, here we compare models with cross-layer parameter sharing and model width reduction.
% The results are shown in the second row of \fig{fig:embedding_encdec}.

\textbf{Cross-Layer Parameter Sharing (Sharing).}
% Parameter sharing controls the complexity of deep neural networks by forcing different architectures to share the same parameters. 
The most widespread use of parameter sharing is in convolutional neural networks \cite{DBLP:conf/cvpr/LongSD15}. 
In recent years, it has also been investigated on NLP and NLU tasks. 
% For example, Universal Transformer shares parameters recurrently in a self-attentive model \cite{DBLP:conf/iclr/DehghaniGVUK19}. 
% \citet{DBLP:journals/corr/abs-2104-06022} proposes several parameter sharing strategies and demonstrates their effectiveness. 
% ALBERT combines cross-layer parameter sharing and decomposed embedding to compress the BERT-based models \cite{DBLP:conf/iclr/LanCGGSS20}.
Among them, cross-layer parameter sharing can provide stronger nonlinearity along the model depth while keeping the parameters unchanged  \cite{DBLP:conf/iclr/DehghaniGVUK19,DBLP:journals/corr/abs-2104-06022,DBLP:conf/iclr/LanCGGSS20}.

\textbf{Reducing Model Width (Width).}
Since model depth has been proven to be important in natural language processing tasks such as machine translation \cite{DBLP:conf/naacl/DevlinCLT19,DBLP:journals/corr/abs-2008-07772,DBLP:journals/corr/abs-2203-00555,DBLP:journals/corr/abs-2008-07772}, here we keep the depth unchanged and reduce the model width.

\textbf{Reducing Model Width is More Efficient and Quantization-Friendly.}
% \textbf{Both methods are about the same in model accuracy.}
% To compare these two encoder/decoder compression methods, we also conduct experiments on three models mentioned in \tab{tab:detailed}. 
In the second row of \fig{fig:embedding_encdec}, these two methods perform nearly the same.
However, \tab{tab:comparison1} shows that there is a large difference in FLOPs and MMI/O, which means reducing model width is much more efficient in computation and memory.
Since it is necessary to quantize these models for greater compression, we further compare the weights and output ranges of the two methods in \fig{fig:encdec_quantization}.
% The left two figures shown weight and output ranges for each layer. 
% The right figure shows the model performance of Post Training Quantization (PTQ) in cross-layer parameter sharing vs. reducing model width.
It can obviously be observed that models with parameter sharing have larger ranges of values for both weight and output, which is not quantization-friendly.
The right figure also verifies this: when we apply post-training quantization (PTQ)  \cite{DBLP:journals/corr/SungSH15,DBLP:conf/nips/BannerNS19,DBLP:conf/iccvw/ChoukrounKYK19} to these two methods, cross-layer parameter sharing performs poorly. 
% Since PTQ can reflect the degree of compatibility of the model architecture with quantization.

\usetikzlibrary{shapes.geometric}
\pgfdeclareplotmark{mystar-small}{
    \node[star,star point ratio=2.25,minimum size=10pt,
    inner sep=0pt,draw=green,solid,fill=green] {};
}
\pgfdeclareplotmark{mystar-tiny}{
    \node[star,star point ratio=2.25,minimum size=7pt,
    inner sep=0pt,draw=red,solid,fill=red] {};
}
\pgfdeclareplotmark{mystar-mini}{
    \node[star,star point ratio=2.25,minimum size=10pt,
    inner sep=0pt,draw=blue,solid,fill=blue] {};
}
\begin{figure}[t!]
\centering
\small
\begin{tikzpicture} 
    \begin{axis}[
        width=1.0\linewidth,
        height=0.75\linewidth,
        yticklabel style={/pgf/number format/fixed,/pgf/number format/precision=1},
        ylabel={BLEU [\%]},
        ylabel near ticks,
        xlabel={\# Params (M)},
        xlabel near ticks,
        enlargelimits=0.1,
        xmajorgrids=true,
        ymajorgrids=true,
        grid style=dashed,
        every tick label/.append style={font=\small},
        label style={font=\small},
        ylabel style={yshift=5pt},
        legend image post style={scale=1},
        legend cell align={left},
        legend pos=south east,
        legend columns=1,
        ]
        \addplot [
            scatter,
            only marks,
            point meta=explicit symbolic,
            scatter/classes={
            base={mark=*,thick,red,mark size=2.5pt},
            vocab={mark=triangle*,thick,lyyblue,mark size=3pt},
            enc_depth={mark=diamond*,thick,lyygreen,mark size=3pt},
            dec_depth={mark=halfdiamond*,thick,lyygreen,mark size=3pt},
            hidden={mark=diamond*,thick,orange,mark size=3pt},
            ffn={mark=halfdiamond*,thick,orange,mark size=3pt},
            back1={mark=square*,lyyblue!7,mark size=40pt},
            back2={mark=square*,lyyred!7,mark size=40pt},
            my-small={mark=mystar-small},
            my-tiny={mark=mystar-tiny},
            my-mini={mark=mystar-mini}
            }
        ] table [meta=label] {
            x y label  
            15.0 23.00 back1
            29.3 24.50 back2
            21.5 24.20 base
            18.7 24.51 vocab
            16.2 24.20 vocab
            13.7 22.97 vocab
            12.4 22.49 vocab
            24.8 24.88 vocab
            27.3 25.02 vocab
            17.5 22.47 enc_depth
            18.3 23.06 enc_depth
            19.9 23.89 enc_depth
            23.0 24.65 enc_depth
            24.6 25.18 enc_depth
            26.1 25.22 enc_depth
            27.8 25.50 enc_depth
            29.3 25.90 enc_depth
            16.2 22.98 dec_depth
            17.2 23.60 dec_depth
            19.3 24.00 dec_depth
            23.6 24.33 dec_depth
            25.7 24.50 dec_depth
            27.8 24.90 dec_depth
            29.9 24.58 dec_depth
            32.0 25.00 dec_depth
            35.7 25.28 hidden
            9.5 21.57 hidden
            15.2 23.43 hidden
            28.3 25.13 hidden
            18.3 23.51 ffn
            24.6 24.75 ffn
            16.7 23.34 ffn
            27.8 24.97 ffn
        };
        \addlegendentry{Small Baseline}
        \addlegendentry{Vocab Size}
        \addlegendentry{Encoder Depth}
        \addlegendentry{Decoder Depth}
        \addlegendentry{Hidden Size}
        \addlegendentry{FFN Dim}
        \end{axis}
    \draw [dashed, thick, red!80] (2.8,0.4) -- (0.3,2.9);
    \draw[color=red, thick] (2.8,0.4) circle (0.3);
    \draw[color=red, thick] (0.3,2.9) circle (0.12);
    % \node at (2.8,0.4) {more important};
    % \node at (0.4,2.9) {less important};
    \draw [dashed, thick, red!80] (5.5,2.8) -- (3.1,3.8);
    \draw[color=red, thick] (3.1,3.8) circle (0.3);
    \draw[color=red, thick] (5.5,2.82) circle (0.12);
    % \draw[very thick,gray!50] (1.9,1.5)..controls (5,7)and(8,7.5) .. (11.5,7.3);
    % \draw[very thick,gray!50] (2.1,1.2)..controls (5,7)and(8,7.0) .. (12.5,7.0);
\end{tikzpicture}
\hspace{\fill}
\caption{Performance (BLEU) vs. parameters (M). Different marks denote different dimensions. Points near large red circles have a greater impact on model performance than points near small red circles. \uwave{Encoder depth can be considered as the most important dimension.}} 
\label{fig:efficiency}
\end{figure}
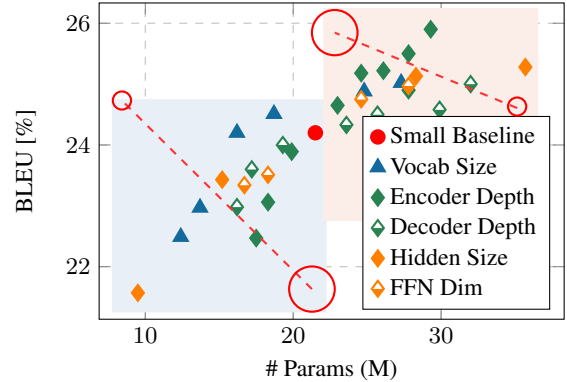

\subsection{Deep Encoder and Shallow Decoder}
\label{subsec:tradeoff}

\fig{fig:efficiency} studies how different dimensions affect the Transformer performance.
% Here we select the small model with 21.5M parameters in \tab{tab:detailed} to explore the model behavior when parameters are limited.
In order to analyze the impact of each dimension separately, here we only change one specific dimension and keep the others unchanged.
The point {\colorbox{lyyblue!15}{on the left}} of the Small Baseline {\color{red}{$\CIRCLE$}} represents scaling one dimension down, while the point {\colorbox{lyyred!15}{on the right}} represents scaling one dimension up.
% In the left part of the figure, the points on the top left indicate that changing this dimension has less impact on model performance, and points closer to the bottom right indicate that changing this dimension has a greater impact.
% The situation in the right is the opposite of that in the left.
We can see that Encoder Depth {\color{lyygreen}{$\blacklozenge$}} is more important than other dimensions, which is consistent with the related work on large-scale models \cite{DBLP:conf/acl/WangLXZLWC19,DBLP:journals/corr/abs-2203-00555}.
% It can also be seen that Vocab Size {\color{lyyblue}{$\blacktriangle$}} is less important.
Based on the above discussion, we finally build a deep encoder and a shallow decoder, while reducing the vocab size and model width.
Two MobileNMT models of different sizes are built here and the detailed settings are shown in \tab{tab:detailed_mobilenmt}.
% Two MobileNMT models of different sizes, named MobileNMT-10MB and MobileNMT-20MB, are built here. 
% \tab{tab:architecture} summarizes the results of adding each part from this section.
% The model size and decoding speed have been significantly improved by redesigning the architecture.
% The degraded BLEU will be improved by Section \ref{sec:training}.

\section{Training Strategies}
\label{sec:training}

\begin{table}[t!]
  \centering
  \small
  \renewcommand\arraystretch{1.0}
  \setlength{\tabcolsep}{0.4mm}
  \begin{tabular}{c|c|cccc|cccc}
      \hline
      \multicolumn{1}{c|}{\multirow{1}{*}{Module}} & Dim & \multicolumn{4}{c|}{MobileNMT-10MB} & \multicolumn{4}{c}{MobileNMT-20MB} \\
      \hline
      \multicolumn{1}{c|}{\multirow{3}{*}{Embed}} & Vocab & \multicolumn{1}{r}{\multirow{3}{*}{\Huge{$\left[\right.$}}} & 8,000 & \multirow{3}{*}{{\Huge$\left.\right]$}} & & \multicolumn{1}{r}{\multirow{3}{*}{\Huge{$\left[\right.$}}} & 8,000 & \multirow{3}{*}{{\Huge$\left.\right]$}} \\
      \multicolumn{1}{c|}{\multirow{1}{*}{}} & Embed & & N/A & & $\times1$ & & N/A & & $\times1$\\
      \multicolumn{1}{c|}{\multirow{1}{*}{}} & Hidden & & 256 & & & & 384 & &\\
      \hline
      \multicolumn{1}{c|}{\multirow{3}{*}{Encoder}} & Hidden & \multicolumn{1}{r}{\multirow{3}{*}{\Huge{$\left[\right.$}}} & 256 & \multirow{3}{*}{{\Huge$\left.\right]$}} & & \multicolumn{1}{r}{\multirow{3}{*}{\Huge{$\left[\right.$}}} & 384 & \multirow{3}{*}{{\Huge$\left.\right]$}} & \\
      \multicolumn{1}{c|}{\multirow{1}{*}{}} & Head & & 4 & & $\times12$ & & 6 & & $\times12$ \\
      \multicolumn{1}{c|}{\multirow{1}{*}{}} & FFN & & 512 & & & & 768 & & \\
      \hline
      \multicolumn{1}{c|}{\multirow{3}{*}{Decoder}} & Hidden & \multicolumn{1}{r}{\multirow{3}{*}{\Huge{$\left[\right.$}}} & 256 & \multirow{3}{*}{{\Huge$\left.\right]$}} & & \multicolumn{1}{r}{\multirow{3}{*}{\Huge{$\left[\right.$}}} & 384 & \multirow{3}{*}{{\Huge$\left.\right]$}} & \\
      \multicolumn{1}{c|}{\multirow{1}{*}{}} & Head & & 4 & & $\times2$ & & 6 & & $\times2$ \\
      \multicolumn{1}{c|}{\multirow{1}{*}{}} & FFN & & 512 & & & & 768 & & \\
      \hline
      \multicolumn{2}{c|}{\multirow{1}{*}{Params}} & \multicolumn{4}{c|}{\multirow{1}{*}{$\approx$10M}} & \multicolumn{4}{c}{\multirow{1}{*}{$\approx$20M}} \\
      \hline
  \end{tabular}
  \caption{The detailed settings of MobileNMT.}
  \label{tab:detailed_mobilenmt}
\end{table}

\begin{figure}[t!]
\centering
\small
% \hspace*{\fill}
\subfigure[FFN Quantizer]
{
   \begin{tikzpicture}
   \tikzstyle{opnode} = [rectangle,draw,rounded corners=2pt,minimum width=0.9\base,minimum height=0.4\base,font=\small,align=center,inner sep=0pt]
   \tikzstyle{labelnode} = [font=\small,inner sep=0pt]

   \baseh=0.41\base
   \basew=0.4\base

   \begin{scope}
       \node [font=\small,inner sep=0pt] (input) at (0,0) {$X$};
       \node [opnode,anchor=south,fill=blue!30!white] (ln) at ([yshift=\baseh]input.north) {LN};
       \node [anchor=south,inner sep=0pt] (matmul1) at ([yshift=\baseh]ln.north) {$\bigotimes$};
       \node [anchor=south,inner sep=0pt] (add1) at ([yshift=\baseh]matmul1.north) {$\bigoplus$};
       \node [opnode,anchor=south,fill=ugreen!40!white] (relu) at ([yshift=\baseh]add1.north) {\relu{}};
       \node [anchor=south,inner sep=0pt] (matmul2) at ([yshift=\baseh]relu.north) {$\bigotimes$};
       \node [anchor=south,inner sep=0pt] (add2) at ([yshift=\baseh]matmul2.north) {$\bigoplus$};
       \node [anchor=south,inner sep=0pt] (add) at ([yshift=\baseh]add2.north) {$\bigoplus$};
       \node [font=\small,inner sep=0pt,anchor=south] (output) at ([yshift=0.3\base]add.north) {$Y_f$};

       \draw [-latex'] (input) to (ln);
       \draw [-latex',red,thick] (ln) to (matmul1);
       \draw [-latex', red,thick] (matmul1) to (add1);
       \draw [-latex'] (add1) to (relu);
       \draw [-latex', red,thick] (relu) to (matmul2);
       \draw [-latex', red,thick] (matmul2) to (add2);
       \draw [-latex'] (add2) to (add);
       \draw [-latex'] (add) to (output);

       \node [font=\small,anchor=east] (w1) at ([xshift=-\basew]matmul1.west) {$W_1$};
       \node [font=\small,anchor=east] (b1) at ([xshift=-\basew]add1.west) {$b_1$};
       \node [font=\small,anchor=east] (w2) at ([xshift=-\basew]matmul2.west) {$W_2$};
       \node [font=\small,anchor=east] (b2) at ([xshift=-\basew]add2.west) {$b_2$};

       \draw [-latex', red,thick] (w1) to (matmul1);
       \draw [-latex'] (b1) to (add1);
       \draw [-latex', red,thick] (w2) to (matmul2);
       \draw [-latex'] (b2) to (add2);

       \coordinate (corner) at ([xshift=0.6\base]ln.east);
       \node [opnode,fill=red!30!white] (gamma) at ([xshift=0.9\base]add2.east) {$\gamma_i$};
       \draw [-latex',rounded corners=5pt] (input.north) to (corner) to (gamma.south);\begin{pgfonlayer}{background}
       \draw [-latex',rounded corners=5pt] (gamma.south) to (gamma.north) to (add.east);
       \end{pgfonlayer}
   \end{scope}
   \end{tikzpicture}
   \label{fig:ffn_quantizer}
}
% \hfill
\subfigure[Attention Quantizer]
{
   \begin{tikzpicture}
   \tikzstyle{opnode} = [rectangle,draw,rounded corners=2pt,minimum width=0.9\base,minimum height=0.4\base,font=\small,align=center,inner sep=0pt]
   \tikzstyle{labelnode} = [font=\small,inner sep=0pt]

   \baseh=0.2\base
   \basew=0.4\base

   \begin{scope}
       \node [font=\small,inner sep=0pt] (input) at (0,0) {$X$};
       \node [opnode,anchor=south,fill=blue!30!white] (ln) at ([yshift=\baseh]input.north) {LN};
       \node [anchor=south,inner sep=0pt] (matmul1) at ([yshift=\baseh]ln.north) {$\bigotimes$};
       \node [anchor=south,inner sep=0pt] (add1) at ([yshift=\baseh]matmul1.north) {$\bigoplus$};
       \node [anchor=south,inner sep=0pt] (K) at ([yshift=\baseh]add1.north) {$K$};
       \node [anchor=south,inner sep=0pt] (matmul3) at ([yshift=\baseh,xshift=-0.65\base]K.north) {$\bigotimes$};
       \node [anchor=south,inner sep=0pt] (matmul4) at ([yshift=\baseh]matmul3.north) {$\bigotimes$};
       \node [opnode,anchor=south,fill=orange!30!white] (softmax) at ([yshift=\baseh]matmul4.north) {\softmax{}};
       \node [anchor=south,inner sep=0pt] (matmul2) at ([yshift=\baseh,xshift=1.67\base]softmax.north) {$\bigotimes$};
       \node [anchor=south,inner sep=0pt] (add2) at ([yshift=\baseh]matmul2.north) {$\bigoplus$};
       \node [anchor=south,inner sep=0pt] (add) at ([yshift=\baseh]add2.north) {$\bigoplus$};
       \node [font=\small,inner sep=0pt,anchor=south] (output) at ([yshift=0.3\base]add.north) {$Y_a$};
       \node [font=\small,anchor=east] (Q) at ([xshift=-0.8\base]K.west) {$Q$};
       \node [font=\small,anchor=east] (V) at ([xshift=1.1\base]K.east) {$V$};
       \node [anchor=south,inner sep=0pt] (matmul5) at ([yshift=1.2\base]V.north) {$\bigotimes$};

       \draw [-latex'] (input) to (ln);
       \draw [-latex', red,thick] (ln) to (matmul1);
       \draw [-latex', red,thick] (matmul1) to (add1);
       \draw [-latex', red,thick] (add1) to (K);
       \draw [-latex', red,thick] (add1) to (Q);
       \draw [-latex', red,thick] (add1) to (V);
       \draw [-latex', red,thick] (K) to (matmul3);
       \draw [-latex', red,thick] (Q) to (matmul3);
       \draw [-latex'] (matmul3) to (matmul4);
       \draw [-latex'] (matmul4) to (softmax);
       \draw [-latex'] (softmax.east) to (matmul5.west);
       \draw [-latex', red,thick] (V) to (matmul5);
       \draw [-latex', red,thick] (matmul5) to (matmul2);
       \draw [-latex', red,thick] (matmul2) to (add2);
       \draw [-latex'] (add2) to (add);
       \draw [-latex'] (add) to (output);

       \node [font=\small,anchor=east] (w1) at ([xshift=-\basew]matmul1.west) {$W_{qkv}$};
       \node [font=\small,anchor=east] (b1) at ([xshift=-\basew]add1.west) {$b_{qkv}$};
       \node [font=\small,anchor=east] (w2) at ([xshift=-\basew]matmul2.west) {$W_o$};
       \node [font=\small,anchor=east] (b2) at ([xshift=-\basew]add2.west) {$b_o$};
       \node [font=\small,anchor=east] (scale) at ([xshift=-\basew]matmul4.west) {$scale$};

       \draw [-latex', red,thick] (w1) to (matmul1);
       \draw [-latex'] (b1) to (add1);
       \draw [-latex', red,thick] (w2) to (matmul2);
       \draw [-latex'] (b2) to (add2);
       \draw [-latex'] (scale) to (matmul4);

       \coordinate (corner) at ([xshift=1.3\base]ln.east);
       \node [opnode,fill=red!30!white] (gamma) at ([xshift=0.6\base]add2.east) {$\gamma_i$};
       \draw [-latex',rounded corners=5pt] (input.north) to (corner) to (gamma.south);\begin{pgfonlayer}{background}
       \draw [-latex',rounded corners=5pt] (gamma.south) to (gamma.north) to (add.east);
       \end{pgfonlayer}
   \end{scope}
   \end{tikzpicture}
   \label{fig:quantizer-attn}
}
% \hspace*{\fill}
\caption{Running examples of the FFN and attention quantizers. Here red lines denote values that will be quantized, black lines denote values with full precision.}
\label{fig:quantizer}
\end{figure}
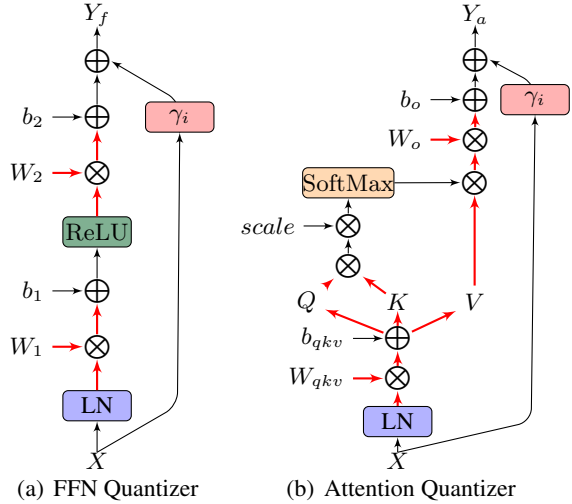

\subsection{Pre-Training with Knowledge Distillation}
\label{subsec:pre-training}

% Knowledge Distillation (KD) is a popular approach to model compression. 
% Recent researchers have tried to combine distillation and quantization to improve the performance of compressed models.
In order to improve the performance of compressed models, recent studies distill knowledge from a well-trained full-precision teacher network to a student network \cite{DBLP:conf/iclr/MishraM18} or directly use a quantized teacher network \cite{DBLP:journals/corr/abs-1911-12491}.
% 111
Here we adopt sequence-level knowledge distillation because it has shown to be effective for NMT tasks.
% and very easy to implement.
The most basic full-precision Transformer-base model is adopted as the teacher.
% Specifically, we use the teacher model to generate pseudo parallel corpus and merge the pseudo corpus with the bilingual parallel corpus to teach the student model. 
% Since the teacher model is only used to construct pseudo corpus, it has no architectural limitations, and 
% here we use the most basic Transformer-base model as the teacher model.

\subsection{Quantization}
\label{subsec:quantization}

The process of quantizing a transformer model can be divided into two steps: 1) constructing quantizers; 2) applying the quantization-aware training (QAT) \cite{DBLP:conf/nips/CourbariauxBD15} based on the pre-trained model we have obtained in Section \ref{subsec:pre-training}.

% \textbf{PTQ and QAT.}
% Quantization enables the model to use lower-bit numbers (such as 8-bit integer) to compute faster and consume less storage space \cite{DBLP:conf/nips/HubaraCSEB16,DBLP:conf/iclr/MicikeviciusNAD18,DBLP:conf/naacl/QuinnB18,DBLP:conf/cvpr/JacobKCZTHAK18}.
% Post-Training Quantization (PTQ) can be seen as the basis for Quantization Aware Training (QAT), it adds quantization nodes to a well-trained floating-point model.
% To quantize a floating-point tensor $r$ to a tensor with $n$ bits, a scale $s$ is introduced to map these two types of values \cite{DBLP:journals/corr/abs-2001-00926}:
% \begin{align}
%     s&=\frac{\max(r)-\min(r)}{2^n-1}
%     \label{eqn:scale}
% \end{align}

% To get a faster computation speed, both weights and activations will be quantized to $n$-bit.
% Suppose $r_m=\min(r)$, the quantization function is:
% \begin{align}
%     Q(r)=\left\lfloor(r-r_m)/s\right\rceil\times s+r_m
%     \label{eqn:quant}
% \end{align}
% where $\left\lfloor \cdot \right\rceil$ represents rounding to the nearest integer.

% However, in PTQ, applying quantization directly to the floating-point network will result in significant performance losses.
% Based on PTQ, QAT simulates the behavior of $n$-bit computation by minimizing quantization errors during training, which helps the model achieve higher accuracy.
% In addition to the learnable weights of the model itself, $s$ is also learnable.

\textbf{FFN and Attention Quantizers.}
% Our MobileNMT is built on the Transformer model \cite{DBLP:conf/nips/VaswaniSPUJGKP17}.
The original Transformer layer includes two types of sublayers: the attention sublayer and feed-forward network (FFN) \cite{DBLP:conf/nips/VaswaniSPUJGKP17}.
Here we construct the quantizer for each linear in the attention and FFN, and quantize both the weights and activations as shown in \fig{fig:quantizer}.
% For the attention sub-layer, we quantize its $W_{q},W_{k},W_{v}$ and $W_{o}$.
% For the FFN sub-layer, we quantize its $W_{1}$ and $W_{2}$.
Since most computations are spent on matrix multiplication, all biases and residuals are kept in full precision for accuracy preservation.
Since quantization will change the range of network outputs, here we add a learnable weight $\gamma_i$ to the $i$-th sublayer to learn how to combine the output and the residual surrounding it.

\begin{figure}[t!]
  \hspace{0.93cm}
  \tikz {
      \small
      \legendmargin=0.03\columnwidth
      \legendwidth=0.08\columnwidth,
      \legendsep=0.025\columnwidth
      \coordinate (start) at (0,0);
      \draw[lyyred,thick,postaction={decorate},decoration={markings,mark=at position 0.5 with {\pgfuseplotmark{*}}}] ([xshift=\legendmargin]start.east) -- +(\legendwidth,0) node[black,right] (l1) {Base};
      \draw[lyyblue,thick,postaction={decorate},decoration={markings,mark=at position 0.5 with {\pgfuseplotmark{square*}}}] ([xshift=\legendsep]l1.east) -- +(\legendwidth,0) node[black,right] (l2) {Ours};
      \draw[lyyred,thick,postaction={decorate},decoration={markings,mark=at position 0.5 with {\pgfuseplotmark{square*}}}] ([xshift=0.1\columnwidth]l2.east) -- +(0.02\columnwidth,0) node[black,right] (l3) {w/o};
      \draw[lyyblue,thick,postaction={decorate},decoration={markings,mark=at position 0.5 with {\pgfuseplotmark{square*}}}] ([xshift=\legendsep]l3.east) -- +(0.02\columnwidth,0) node[black,right] (l4) {w/ PTQ};
      \coordinate (end) at ([xshift=\legendmargin+0pt]l2.east);
      \begin{pgfonlayer}{background}
      \node[rectangle,draw,inner sep=0.2pt] [fit = (start) (l1) (l2) (l3) (l4) (end)] {};
      \end{pgfonlayer}
  }
  \\
  \centering
  \small
  % \hspace*{\fill}
  \begin{tikzpicture}
    \begin{axis}[
      width=0.55\columnwidth,height=0.38\columnwidth,
      yticklabel style={/pgf/number format/fixed,/pgf/number format/precision=1},
      ylabel={BLEU [\%]},
      ylabel near ticks,
      xlabel={Dropout},
      xlabel near ticks,
      enlargelimits=0.1,
      symbolic x coords={0.0, 0.05, 0.1},
      xmajorgrids=true,
      ymajorgrids=true,
      grid style=dashed,
      xtick=data,
      every tick label/.append style={font=\small},
      legend columns=1,
      legend pos=south east,
      legend style={font=\small},
    ]
      \addplot [lyyred,thick,mark=*] coordinates {
        (0.0,24.91) (0.05,27.42) (0.1,27.40) 
      };
      \addplot [lyyblue,thick,mark=square*] coordinates {
        (0.0,25.23) (0.05,24.53) (0.1,23.49) 
      };
    \end{axis}
  \end{tikzpicture}
  \hspace{0.1cm}
  \begin{tikzpicture}
    \begin{axis}[
      width=0.6\columnwidth,height=0.4\columnwidth,
      ybar, ymin=14,ymax=26.6,
      grid style=dashed,
      ymajorgrids=true,
      xmajorgrids=true,
      ylabel near ticks,
      xlabel near ticks,
      enlarge x limits=0.8,
      bar width=8pt,
      symbolic x coords={LR:0.01,LR:0.02},
      xtick=data,
      nodes near coords style={font=\small},
      legend columns=1,
      legend pos=north east,
      legend style={font=\small},
      ]
      \addplot[draw=lyyred,fill=lyyred!90] coordinates {
      (LR:0.01,25.23)(LR:0.02,25.90)};
      \addplot[draw=lyyblue,fill=lyyblue!90] coordinates {
      (LR:0.01,23.05)(LR:0.02,14.80)};
    \end{axis}
  \end{tikzpicture}
  \caption{The left part shows performance of different dropouts on base model vs. MobileNMT. The right part shows performance before vs. after PTQ. \uwave{Removing dropout from MobileNMT can lead to significant performance improvement. While larger learning rates can also improve model performance, the model will become quantization-unfriendly.}}
  % (Increasing the learning rate from 1$\times 10^{-3}$ to 2$\times 10^{-3}$ represents adopting the deep configuration)
  \label{fig:dropout_config}
\end{figure}
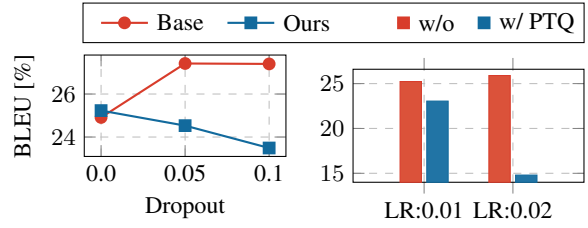

\begin{figure}[t!]
\hspace{0.75cm}
  \tikz {
      \small
      \legendmargin=0.14\columnwidth
      \legendwidth=0.1\columnwidth
      \legendsep=0.12\columnwidth
      \coordinate (start) at (0,0);
      \draw[lyyred,very thick] ([xshift=\legendmargin]start.east) -- +(0.03\columnwidth,0) node[black,right] (l1) {LR: 0.01};
      \draw[lyyblue,very thick] ([xshift=\legendsep]l1.east) -- +(0.03\columnwidth,0) node[black,right] (l2) {LR: 0.02};
      \coordinate (end) at ([xshift=\legendmargin+0pt]l2.east);
      \begin{pgfonlayer}{background}
      \node[rectangle,draw,inner sep=0.2pt] [fit = (start) (l1) (l2) (end)] {};
      \end{pgfonlayer}
  }
  \\[2.5pt]
  \centering
  \small
  % \hspace*{\fill}
  \begin{tikzpicture}
    \begin{axis}[
      width=0.52\columnwidth,height=0.37\columnwidth,
      yticklabel style={/pgf/number format/fixed,/pgf/number format/precision=1},
      ylabel={Weight},
      ylabel near ticks,
      xlabel={\# Layer ID},
      xlabel near ticks,
      enlargelimits=0.1,
      % symbolic x coords={45,50,55,60,65},
      xmajorgrids=true,
      ymajorgrids=true,
      grid style=dashed,
      xtick distance=2,
      every tick label/.append style={font=\small},
      legend columns=1,
      legend pos=north east,
      legend style={font=\small},
    ]
      \addplot [lyyred,very thick] coordinates {
        (0,2.06) (1,1.61) (2,1.49) (3,0.69) (4,0.97) (5,1.99) (6,1.54) (7,2.08) (8,0.99) (9,0.81) (10,1.89) (11,0.95) (12,1.17) (13,1.45)
      };
      \addplot [lyyblue,very thick] coordinates {
        (0,5.51) (1,4.05) (2,4.66) (3,2.56) (4,2.68) (5,2.62) (6,3.49) (7,3.00) (8,2.76) (9,2.82) (10,2.42) (11,3.28) (12,2.01) (13,4.56)
      };
    \end{axis}
  \end{tikzpicture}
  \begin{tikzpicture}
    \begin{axis}[
      width=0.52\columnwidth,height=0.37\columnwidth,
      yticklabel style={/pgf/number format/fixed,/pgf/number format/precision=1},
      ylabel={Output},
      ylabel near ticks,
      xlabel={\# Layer ID},
      xlabel near ticks,
      enlargelimits=0.1,
      % symbolic x coords={45,50,55,60,65},
      xmajorgrids=true,
      ymajorgrids=true,
      grid style=dashed,
      xtick distance=2,
      every tick label/.append style={font=\small},
      legend columns=1,
      legend pos=north east,
      legend style={font=\small},
    ]
      \addplot [lyyred,very thick] coordinates {
        (0,14.30) (1,9.45) (2,6.98) (3,5.35) (4,6.51) (5,71.13) (6,14.85) (7,181.5) (8,21.20) (9,12.42) (10,52.42) (11,41.12) (12,40.25) (13,204.5)
      };
      \addplot [lyyblue,very thick] coordinates {
        (0,591.5) (1,49.5) (2,711.5) (3,27.06) (4,49.56) (5,40.84) (6,129.0) (7,172.12) (8,80.87) (9,144.25) (10,102.31) (11,124.43) (12,222.12) (13,503.5)
      };
    \end{axis}
  \end{tikzpicture}
  \caption{Weight and output ranges for each layer. \uwave{Larger learning rate will result in larger range of values.}}
  \label{fig:deep_range}
\end{figure}
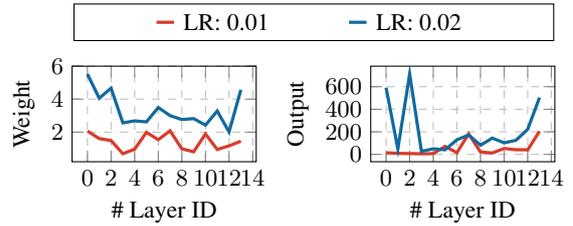

\begin{figure*}[t!]
  \hspace{0.8cm}
  \tikz {
      \small
      \legendmargin=0.04\columnwidth
      \legendwidth=0.074\columnwidth,
      \legendsep=0.02\columnwidth
      \coordinate (start) at (0,0);
      \draw[lyyred,very thick] ([xshift=\legendmargin]start.east) -- +(\legendwidth,0) node[black,right] (l1) {Attn (w/o $L_2$)};
      \draw[lyyred,very thick, dashed] ([xshift=\legendsep]l1.east) -- +(\legendwidth,0) node[black,right] (l2) {Attn (w/ $L_2$)};
      \draw[lyyblue,very thick] ([xshift=\legendsep]l2.east) -- +(\legendwidth,0) node[black,right] (l3) {FFN (w/o $L_2$)};
      \draw[lyyblue,very thick, dashed] ([xshift=\legendsep]l3.east) -- +(\legendwidth,0) node[black,right] (l4) {FFN (w/ $L_2$)};
      \coordinate (end) at ([xshift=\legendmargin+0pt]l4.east);
      \begin{pgfonlayer}{background}
      \node[rectangle,draw,inner sep=0.2pt] [fit = (start) (l1) (l2) (l3) (l4) (end)] {};
      \end{pgfonlayer}
  }
  \hspace{0.92cm}
  \tikz {
      \small
      \legendmargin=0.025\columnwidth
      \legendwidth=0.01\columnwidth,
      \legendsep=0.02\columnwidth
      \coordinate (start) at (0,0);
      \draw[lyyred,thick,postaction={decorate},decoration={markings,mark=at position 0.5 with {\pgfuseplotmark{square*}}}] ([xshift=\legendmargin]start.east) -- +(\legendwidth,0) node[black,right] (l1) {w/o};
      \draw[lyyblue,thick,postaction={decorate},decoration={markings,mark=at position 0.5 with {\pgfuseplotmark{square*}}}] ([xshift=\legendsep]l1.east) -- +(\legendwidth,0) node[black,right] (l2) {w/ PTQ};
      \coordinate (end) at ([xshift=\legendmargin+0pt]l2.east);
      \begin{pgfonlayer}{background}
      \node[rectangle,draw,inner sep=0.2pt] [fit = (start) (l1) (l2) (end)] {};
      \end{pgfonlayer}
  }
  \\[1.5pt]
  \centering
  \small
  \begin{tikzpicture}
    \begin{axis}[
      width=0.85\columnwidth,height=0.4\columnwidth,
      yticklabel style={/pgf/number format/fixed,/pgf/number format/precision=1},
      ylabel={Weight},
      ylabel near ticks,
      xlabel={\# Layer ID},
      xlabel near ticks,
      enlargelimits=0.1,
      % symbolic x coords={45,50,55,60,65},
      xmajorgrids=true,
      ymajorgrids=true,
      grid style=dashed,
      xtick=data,
      every tick label/.append style={font=\small},
      legend columns=1,
      legend pos=north east,
      legend style={font=\small},
    ]
      \addplot [lyyred,very thick] coordinates {
        (0,2.22) (1,3.71) (2,2.47) (3,1.97) (4,1.79) (5,2.01) (6,2.48) (7,2.32) (8,2.29) (9,2.95) (10,2.61) (11,2.38) (12,1.91) (13,2.53)
      };
      \addplot [lyyred,very thick, dashed] coordinates {
        (0,0.38) (1,0.37) (2,0.38) (3,0.39) (4,0.41) (5,0.43) (6,0.73) (7,0.65) (8,0.55) (9,0.98) (10,0.98) (11,0.67) (12,0.67) (13,1.03)
      };
      \addplot [lyyblue,very thick] coordinates {
        (0,5.51) (1,4.05) (2,4.66) (3,2.56) (4,2.68) (5,2.62) (6,3.49) (7,3.00) (8,2.76) (9,2.82) (10,2.42) (11,3.28) (12,2.01) (13,4.56)
      };
      \addplot [lyyblue,very thick, dashed] coordinates {
        (0,1.53) (1,1.07) (2,1.23) (3,0.53) (4,0.75) (5,0.54) (6,0.59) (7,0.51) (8,0.76) (9,0.96) (10,0.66) (11,0.76) (12,1.22) (13,2.01)
      };
    \end{axis}
  \end{tikzpicture}
  \begin{tikzpicture}
    \begin{axis}[
      width=0.85\columnwidth,height=0.4\columnwidth,
      yticklabel style={/pgf/number format/fixed,/pgf/number format/precision=1},
      ylabel={Output},
      ylabel near ticks,
      xlabel={\# Layer ID},
      xlabel near ticks,
      enlargelimits=0.1,
      % symbolic x coords={45,50,55,60,65},
      xmajorgrids=true,
      ymajorgrids=true,
      grid style=dashed,
      xtick=data,
      every tick label/.append style={font=\small},
      legend columns=1,
      legend pos=north east,
      legend style={font=\small},
    ]
      \addplot [lyyred,very thick] coordinates {
        (0,30.25) (1,225.0) (2,15.78) (3,19.40) (4,30.87) (5,18.62) (6,93.37) (7,30.10) (8,91.5) (9,108.25) (10,69.93) (11,28.70) (12,31.31) (13,199.0)
      };
      \addplot [lyyred,very thick, dashed] coordinates {
        (0,5.37) (1,2.47) (2,2.42) (3,2.36) (4,1.87) (5,2.10) (6,8.36) (7,3.61) (8,4.43) (9,5.65) (10,14.89) (11,3.94) (12,4.83) (13,15.75)
      };
      \addplot [lyyblue,very thick] coordinates {
        (0,591.5) (1,49.5) (2,711.5) (3,27.06) (4,49.56) (5,40.84) (6,129.0) (7,172.12) (8,80.87) (9,144.25) (10,102.31) (11,124.43) (12,222.12) (13,503.5)
      };
      \addplot [lyyblue,very thick, dashed] coordinates {
        (0,227.75) (1,4.71) (2,13.67) (3,2.00) (4,2.13) (5,2.05) (6,3.52) (7,2.86) (8,3.43) (9,4.48) (10,8.10) (11,6.92) (12,15.85) (13,42.0)
      };
    \end{axis}
  \end{tikzpicture}
  \begin{tikzpicture}
    \begin{axis}[
      width=0.55\columnwidth,height=0.38\columnwidth,
      ybar, ymin=13,ymax=26.5,
      grid style=dashed,
      ymajorgrids=true,
      xmajorgrids=true,
      ylabel={BLEU [\%]},
      xlabel={Weight Decay},
      ylabel near ticks,
      xlabel near ticks,
      enlarge x limits=0.2,
      bar width=8pt,
      symbolic x coords={0.0,0.05,0.1},
      xtick=data,
      nodes near coords style={font=\small},
      legend columns=1,
      legend pos=north east,
      legend style={font=\small},
      ]
      \addplot[draw=lyyred,fill=lyyred!90] coordinates {
      (0.0,25.90)(0.05,25.79)(0.1,25.35)};
      \addplot[draw=lyyblue,fill=lyyblue!90] coordinates {
      (0.0,14.80)(0.05,23.56)(0.1,21.87)};
    \end{axis}
  \end{tikzpicture}
  \caption{The left two figures show weight and output ranges for each layer. The right figure shows the performance of different $L_2$ regularizations before vs. after PTQ. \uwave{Experiments show that $L_2$ regularization can make the model more quantization-friendly.}}
  \label{fig:l2}
\end{figure*}
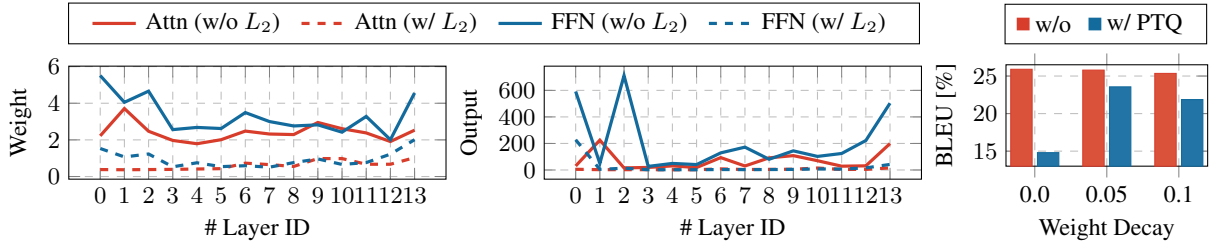

\textbf{Quantization-Aware Training.}
Since MobileNMT only has 10M/20M parameters, quantizing such a small model inevitably results in performance loss, so we perform QAT after constructing the quantizers.
Before QAT, we pre-compute all scaling parameters based on a forward running on the pre-trained distillation model obtained in Section \ref{subsec:pre-training}.
It takes nearly no additional costs, but provides a good initialization.
% At the pre-computation stage, we first clip each activation by a factor $\alpha$ but keep weights unchanged.
% At the quantization aware training stage, we first initialize the network with the pre-trained distillation model and the computed scaling values, then quantize both weights and activations to $n$-bit.
% Then we will pre-compute all scaling parameters based on a forward running on the validation set.
For engineering development, we choose the uniform quantization scheme because of it is hardware-friendly \cite{DBLP:conf/cvpr/LiuCHXS22}. 
% In this work, we quantize both weights and activations. 
For 8-bit quantization, we use the element-wise quantization \cite{DBLP:conf/cvpr/Lee0H21}.
For lower-bit quantization, such as 4-bit integer, we use the row-wise quantization \cite{DBLP:conf/cvpr/FaraoneFBL18}.

% Although the scaling parameters are learnable, we will pre-compute all scaling parameters based on a forward running on the validation set.
% It takes nearly no additional costs, but provides a good initialization for QAT and has been shown to be effective in our experiments.

% 111

% because the distributions of weights are more concentrated than activations.
% To avoid clipping in extreme cases, here we will set a threshold $t$ to limit the minimum value after clipping:
% \begin{align}
%     r_c&=\mathrm{Clip}(r, \alpha, t)
% \end{align}
% where $r$ is the network input, $t=0.3\max(|r|)$. 

% and has been shown to be effective in our experiments.
% Here we can rewrite \eqn{eqn:scale} as:
% \begin{align}
%     s&=\frac{\max(r_c)-\min(r_c)}{2^n-1}
% \end{align}
% where $s$ is the computed scaling value. In the element-wise quantization scheme, $s$ is a single floating-point value.
% In the row-wise quantization scheme, $\max(r_c)$ and $\min(r_c)$ are performed along the first dimension of $r$, producing a scaling value with the same shape as $r$.

% Suppose $r_m=\min(r_c)$, the quantization function in \eqn{eqn:quant} can be rewrited as:
% \begin{align}
%     Q(r_c)=\left\lfloor(r_c-r_m)/s\right\rceil\times s+r_m
% \end{align}
% where $\left\lfloor \cdot \right\rceil$ represents rounding to the nearest integer.

% \textbf{Scaling Distribution.}

\subsection{Training Hyperparameters}

Compared to the original Transformer model, MobileNMT introduced in Section \ref{sec:architecture} has fewer parameters and different architectures, so different training hyperparameters are needed.

\textbf{Removing Dropout.}
Since our models have fewer parameters, we do not need to impose strong regularizations on them and we remove dropout from the entire model.
The left part of \fig{fig:dropout_config} shows that removing dropout will lead to an improvement of almost two BLEU points.

\textbf{Larger Learning Rate.}
Here we follow the configuration provided in \citet{DBLP:conf/acl/WangLXZLWC19} with a larger learning rate ($0.01\rightarrow 0.02$), a larger training batch ($4096\rightarrow 8192$), and more warmup steps ($4000\rightarrow 8000$).
As shown in the right part of \fig{fig:dropout_config}, it can improve model performance by more than 0.5 BLEU points (red bars). 
However, after PTQ, the model with $0.02$ learning rate performs significantly worse than $0.01$ (blue bars).
As shown in \fig{fig:deep_range}, the network weights and outputs become larger when using a larger learning rate, which is not quantization-friendly.

% \begin{table}[!t]
%   \centering
%   \small
%   \renewcommand\arraystretch{1.15}
%   \setlength{\tabcolsep}{1.0mm}{
%     \begin{tabular}{l|c|c|c|c}
%       \hline
%       \multicolumn{1}{c|}{\multirow{2}{*}{System}} &
%       \multicolumn{1}{c|}{\multirow{1}{*}{Params}} &
%        \multicolumn{1}{c|}{\multirow{1}{*}{Size}} &
%       \multicolumn{1}{c|}{Speed} &
%       \multicolumn{1}{c}{\multirow{2}{*}{BLEU}} \\
%       % \cline{4-6}
%       & (M) & (MB) & (sent./s) \\
%       \hline
%       \multirow{1}{*}{Transformer-base} & 65 & 260 & - & 27.40 \\
%       \multirow{1}{*}{Transformer-big} & 218 & 872 & - & 28.36 \\
%       \hline
%       \multirow{1}{*}{+ Section \ref{sec:architecture}} & 10 & 40 & - & 22.54 \\
%       \multirow{1}{*}{+ Distillation} & 10 & 40 & - & 23.77 \\
%       \multirow{1}{*}{+ Quantization} & 10 & 10 & - & 23.76 \\
%       \multirow{1}{*}{+ Hyperparameters} & 10 & 10 & - & 25.48 \\
%       \hline
%     \end{tabular}
%   \caption{Results of adding each part from Section \ref{sec:training}.}
%   \label{tab:training}
%   }
% \end{table}

\textbf{$L_2$ Regularization.}
To solve the above problem, this paper adopts $L_2$ regularization applied to weight (also called weight decay).
It adds the squared magnitude of the network weights as the penalty term to the original loss function and encourage the weights to be smaller.
% \begin{align}
%     \mathcal{L}_{\lambda}(w)=\mathcal{L}(w)+\lambda\norm{w}_2^2
% \end{align}
% where $\mathcal{L}$ is the original cross-entropy loss, $\lambda$ is the penalty factor.
% The second $L_2$ penalty term will encourage the network weights $w$ to be smaller.
As shown in the left two parts of \fig{fig:l2}, with $L_2$ regularization, both the network weights and output values will become significantly smaller.
The right part of \fig{fig:l2} shows the performance of PTQ when applying different degrees of $L_2$ regularization.
% When weight decay is set to 0.0, it means that there is no $L_2$ regularization.
The red and blue bars represent the model performance before and after PTQ.
We can see that $L_2$ regularization does improve the model performance after PTQ.
% From these red bars, we can see that $L_2$ regularization will not bring significant performance degradation on models without PTQ.
% However, with PTQ, adding $L_2$ regularization will significantly improve models with deep configuration.

\section{The Engine}
\label{sec:engine}

This section introduces the detailed implementations of our inference engine. 

\begin{table*}[t!]
  \centering
  \small
  \renewcommand\arraystretch{1.1}
  \setlength{\tabcolsep}{0.18cm}
  \begin{tabular}{c|l|l|l|l|l|l|l}
  \hline
  &
  \multicolumn{1}{c|}{\multirow{1}{*}{System}} &
  \multicolumn{1}{c|}{\multirow{1}{*}{Params (M)}} & \multicolumn{1}{c|}{\multirow{1}{*}{Size (MB)}} & \multicolumn{1}{c|}{\multirow{1}{*}{Memory (MB)}} & \multicolumn{1}{c|}{\multirow{1}{*}{Latency (ms)}} & \multicolumn{1}{c|}{\multirow{1}{*}{Test}} & \multicolumn{1}{c}{\multirow{1}{*}{Valid}} \\
  % \cline{6-10}
  % & \multicolumn{1}{c|}{\multirow{1}{*}{(M)}} & \multicolumn{1}{c|}{\multirow{1}{*}{(MB)}} & & & & \multicolumn{1}{c|}{\multirow{1}{*}{(M)}} & \multicolumn{1}{c|}{\multirow{1}{*}{(MB)}} & & & \\
  \hline
  \multirow{6}*{\rotatebox{90}{En-De}} & Transformer-big & 218 {\small{\textcolor{red}{$\uparrow$1$\times$}}} & 872 {\small{\textcolor{red}{$\uparrow$1$\times$}}} & 2886.6 {\small{\textcolor{red}{$\uparrow$1.0$\times$}}} & 1281.5 {\small{\textcolor{red}{$\uparrow$1.0$\times$}}} & 28.36 {\small{\textcolor{blue}{$\Delta$-0.00}}} & 26.75 {\small{\textcolor{blue}{$\Delta$-0.00}}} \\
  & Transformer-base & 65 {\small{\textcolor{red}{$\uparrow$3$\times$}}} & 260 {\small{\textcolor{red}{$\uparrow$3$\times$}}} & 908.5 {\small{\textcolor{red}{$\uparrow$3.2$\times$}}} & 332.3 {\small{\textcolor{red}{$\uparrow$3.9$\times$}}} & 27.40 {\small{\textcolor{blue}{$\Delta$-0.96}}} & 25.81 {\small{\textcolor{blue}{$\Delta$-0.94}}} \\
  & Transformer-small & 22 {\small{\textcolor{red}{$\uparrow$10$\times$}}} & 88 {\small{\textcolor{red}{$\uparrow$10$\times$}}} & 759.5 {\small{\textcolor{red}{$\uparrow$3.8$\times$}}} & 158.0 {\small{\textcolor{red}{$\uparrow$8.1$\times$}}} & 24.20 {\small{\textcolor{blue}{$\Delta$-4.61}}} & 23.91 {\small{\textcolor{blue}{$\Delta$-2.84}}} \\
  & Transformer-tiny & 8 {\small{\textcolor{red}{$\uparrow$27$\times$}}} & 32 {\small{\textcolor{red}{$\uparrow$27$\times$}}} & 398.9 {\small{\textcolor{red}{$\uparrow$7.2$\times$}}} & 73.0 {\small{\textcolor{red}{$\uparrow$17.6$\times$}}} & 20.97 {\small{\textcolor{blue}{$\Delta$-7.39}}} & 21.53 {\small{\textcolor{blue}{$\Delta$-5.22}}} \\
  & MobileNMT-20MB & 20 {\small{\textcolor{red}{$\uparrow$11$\times$}}} & 20 {\small{\textcolor{red}{$\uparrow$44$\times$}}} & 26.0 {\small{\textcolor{red}{$\uparrow$111.2$\times$}}} & 46.3 {\small{\textcolor{red}{$\uparrow$27.7$\times$}}} & 27.09 {\small{\textcolor{blue}{$\Delta$-1.27}}} & 25.72 {\small{\textcolor{blue}{$\Delta$-1.03}}} \\
  & MobileNMT-10MB & 10 {\small{\textcolor{red}{$\uparrow$22$\times$}}} & 10 {\small{\textcolor{red}{$\uparrow$87$\times$}}} & 14.9 {\small{\textcolor{red}{$\uparrow$194.0$\times$}}} & 27.3 {\small{\textcolor{red}{$\uparrow$47.0$\times$}}} & 25.08 {\small{\textcolor{blue}{$\Delta$-3.28}}} & 24.85 {\small{\textcolor{blue}{$\Delta$-1.90}}} \\
  \hline
  \hline
  \multirow{6}*{\rotatebox{90}{En-Fr}} & Transformer-big & 259 {\small{\textcolor{red}{$\uparrow$1$\times$}}} & 1036 {\small{\textcolor{red}{$\uparrow$1$\times$}}} & 2987.6 {\small{\textcolor{red}{$\uparrow$1.0$\times$}}} & 1345.6 {\small{\textcolor{red}{$\uparrow$1.0$\times$}}} & 39.05 {\small{\textcolor{blue}{$\Delta$-0.00}}} & 44.12 {\small{\textcolor{blue}{$\Delta$-0.00}}} \\
  & Transformer-base & 86 {\small{\textcolor{red}{$\uparrow$3$\times$}}} & 344 {\small{\textcolor{red}{$\uparrow$3$\times$}}} & 944.8 {\small{\textcolor{red}{$\uparrow$3.2$\times$}}} & 358.9 {\small{\textcolor{red}{$\uparrow$3.7$\times$}}} & 38.64 {\small{\textcolor{blue}{$\Delta$-0.41}}} & 43.80 {\small{\textcolor{blue}{$\Delta$-0.32}}} \\
  & Transformer-small & 22 {\small{\textcolor{red}{$\uparrow$12$\times$}}} & 88 {\small{\textcolor{red}{$\uparrow$12$\times$}}} & 782.3 {\small{\textcolor{red}{$\uparrow$3.8$\times$}}} & 178.5 {\small{\textcolor{red}{$\uparrow$7.5$\times$}}} & 34.76 {\small{\textcolor{blue}{$\Delta$-4.29}}} & 40.01 {\small{\textcolor{blue}{$\Delta$-4.11}}} \\
  & Transformer-tiny & 8 {\small{\textcolor{red}{$\uparrow$32$\times$}}} & 32 {\small{\textcolor{red}{$\uparrow$32$\times$}}} & 418.8 {\small{\textcolor{red}{$\uparrow$7.1$\times$}}} & 80.3 {\small{\textcolor{red}{$\uparrow$16.8$\times$}}} & 30.36 {\small{\textcolor{blue}{$\Delta$-8.69}}} & 36.01 {\small{\textcolor{blue}{$\Delta$-8.11}}} \\
  & MobileNMT-20MB & 20 {\small{\textcolor{red}{$\uparrow$13$\times$}}} & 20 {\small{\textcolor{red}{$\uparrow$52$\times$}}} & 26.7 {\small{\textcolor{red}{$\uparrow$111.9$\times$}}} & 53.7 {\small{\textcolor{red}{$\uparrow$25.1$\times$}}} & 37.67 {\small{\textcolor{blue}{$\Delta$-1.38}}} & 43.81 {\small{\textcolor{blue}{$\Delta$-0.31}}} \\
  & MobileNMT-10MB & 10 {\small{\textcolor{red}{$\uparrow$26$\times$}}} & 10 {\small{\textcolor{red}{$\uparrow$104$\times$}}} & 15.8 {\small{\textcolor{red}{$\uparrow$189.1$\times$}}} & 28.9 {\small{\textcolor{red}{$\uparrow$46.6$\times$}}} & 36.00 {\small{\textcolor{blue}{$\Delta$-3.05}}} & 41.87 {\small{\textcolor{blue}{$\Delta$-2.25}}} \\
  \hline
  \end{tabular}
  \caption{Results on WMT14 En-De and WMT14 En-Fr tasks. These metrics are measured on Google Pixel 4. Transformer-big/base/small/tiny results are tested on TFLite and MobileNMT-20MB/10MB are tested on our engine. All results are based on a sample with src/tgt length of 30.} 
  % We mark the smallest three models in bold.
  \label{tab:main_result}
\end{table*}

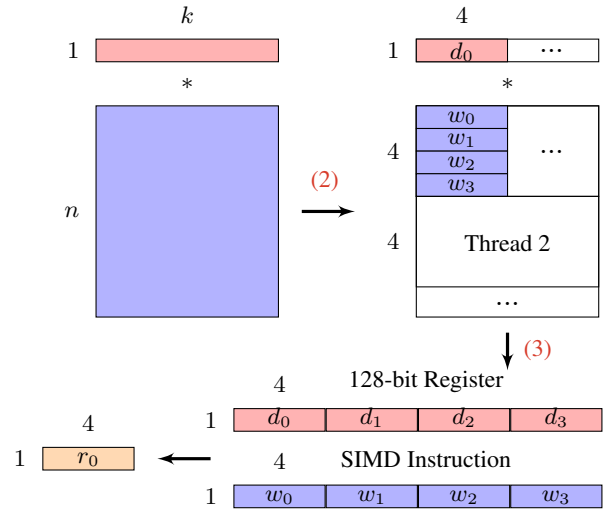
\begin{figure}[t!]
\centering
\small
\begin{tikzpicture}
\tikzstyle{m1} = [rectangle,draw,minimum width=2.4\base,minimum height=0.3\base]
\tikzstyle{m2} = [rectangle,draw,minimum width=2.4\base,minimum height=2.8\base]
\tikzstyle{m3} = [rectangle,draw,minimum width=1.2\base,minimum height=0.3\base]
\tikzstyle{m4} = [rectangle,draw,minimum width=2.4\base,minimum height=1.2\base]
\tikzstyle{m5} = [rectangle,draw,minimum width=1.2\base,minimum height=1.2\base]

\baseh=-1.0\base
\basew=0.4\base

\begin{scope}
   \node [m1,anchor=north,fill=red!30!white] (matrix1) at (0,0){};
   \node [m2,anchor=north,fill=blue!30!white] (matrix2) at ([yshift=\baseh]input.north) {};
   \node [font=\small,anchor=south] () at ([yshift=0.1\base]matrix1.north) {$k$};
   \node [font=\small,anchor=east] () at ([xshift=-0.1\base]matrix1.west) {$1$};
   \node [font=\small,anchor=south] () at ([yshift=0.1\base]matrix2.north) {$*$};
   \node [font=\small,anchor=east] () at ([xshift=-0.1\base]matrix2.west) {$n$};

   \node [m1,anchor=west] (matrix3) at ([xshift=1.8\base]matrix1.east){};
   \node [m3,anchor=west,fill=red!30!white] (matrix31) at ([xshift=1.8\base]matrix1.east){};
   \node [font=\small,anchor=south] () at ([yshift=0.1\base]matrix31.north) {$4$};
   \node [font=\small,anchor=east] () at ([xshift=-0.1\base]matrix31.west) {$1$};
   \node [font=\small,anchor=south] () at ([yshift=-0.1\base]matrix31.south) {$d_0$};
   \node [font=\normalsize,anchor=west] () at ([xshift=0.3\base]matrix31.east) {$...$};
   
   \node [m2,anchor=west] (matrix4) at ([xshift=1.8\base]matrix2.east) {};
   \node [m4,anchor=north] (matrix41) at ([yshift=0.0\base]matrix4.north) {};
   \node [m5,anchor=west,fill=blue!30!white] (matrix411) at ([xshift=0.0\base]matrix41.west) {};
    \node [font=\small,anchor=east] () at ([xshift=-0.1\base]matrix41.west) {$4$};
    \node [font=\small,lyyred,anchor=east] () at ([xshift=-0.9\base,yshift=-0.4\base]matrix41.west) {(2)};
    \node [font=\small,anchor=south] () at ([yshift=0.1\base]matrix41.north) {$*$};
    \node [font=\normalsize,anchor=west] () at ([xshift=0.3\base]matrix411.east) {$...$};

    \node [m3,anchor=north,fill=blue!30!white] (small1) at ([yshift=0.0\base]matrix411.north){};
    \node [font=\small,anchor=south] () at ([yshift=-0.05\base]small1.south) {$w_0$};
    \node [m3,anchor=north,fill=blue!30!white] (small2) at ([yshift=-0.3\base]matrix411.north){};
    \node [font=\small,anchor=south] () at ([yshift=-0.05\base]small2.south) {$w_1$};
    \node [m3,anchor=north,fill=blue!30!white] (small3) at ([yshift=-0.6\base]matrix411.north){};
    \node [font=\small,anchor=south] () at ([yshift=-0.05\base]small3.south) {$w_2$};
    \node [m3,anchor=north,fill=blue!30!white] (small4) at ([yshift=-0.9\base]matrix411.north){};
    \node [font=\small,anchor=south] () at ([yshift=-0.05\base]small4.south) {$w_3$};
   
   \node [m4,anchor=north] (matrix42) at ([yshift=-1.2\base]matrix4.north) {};
   \node [font=\small,anchor=east] () at ([xshift=-0.1\base]matrix42.west) {$4$};
   \node [font=\small,anchor=north] () at ([yshift=-0.4\base]matrix42.north) {Thread 2};
   \node [font=\normalsize,anchor=north] () at ([yshift=-1.3\base]matrix42.north) {$...$};
   \node [font=\small,lyyred,anchor=north] () at ([xshift=0.4\base,yshift=-1.8\base]matrix42.north) {(3)};

   \node [m3,anchor=north,fill=red!30!white] (small11) at ([xshift=-3.0\base,yshift=-1.2\base]matrix4.south){};
   \node [font=\small,anchor=south] () at ([yshift=-0.05\base]small11.south) {$d_0$};
   \node [font=\small,anchor=south] () at ([yshift=0.1\base]small11.north) {$4$};
   \node [font=\small,anchor=east] () at ([xshift=-0.1\base]small11.west) {$1$};
   \node [m3,anchor=west,fill=red!30!white] (small12) at ([yshift=-0.0\base]small11.east){};
   \node [font=\small,anchor=south] () at ([yshift=-0.05\base]small12.south) {$d_1$};
   \node [m3,anchor=west,fill=red!30!white] (small13) at ([yshift=-0.0\base]small12.east){};
   \node [font=\small,anchor=south] () at ([yshift=-0.05\base]small13.south) {$d_2$};
   \node [m3,anchor=west,fill=red!30!white] (small14) at ([yshift=-0.0\base]small13.east){};
   \node [font=\small,anchor=south] () at ([yshift=-0.05\base]small14.south) {$d_3$};
   \node [font=\small,anchor=south] () at ([xshift=-0.5\base,yshift=0.1\base]small13.north) {128-bit Register};
   \node [font=\small,anchor=south] () at ([xshift=-0.5\base,yshift=-0.9\base]small13.north) {SIMD Instruction};

   \node [m3,anchor=north,fill=blue!30!white] (small21) at ([yshift=-0.7\base]small11.south){};
   \node [font=\small,anchor=south] () at ([yshift=-0.05\base]small21.south) {$w_0$};
   \node [font=\small,anchor=south] () at ([yshift=0.1\base]small21.north) {$4$};
   \node [font=\small,anchor=east] () at ([xshift=-0.1\base]small21.west) {$1$};
   \node [m3,anchor=west,fill=blue!30!white] (small22) at ([yshift=-0.0\base]small21.east){};
   \node [font=\small,anchor=south] () at ([yshift=-0.05\base]small22.south) {$w_1$};
   \node [m3,anchor=west,fill=blue!30!white] (small23) at ([yshift=-0.0\base]small22.east){};
   \node [font=\small,anchor=south] () at ([yshift=-0.05\base]small23.south) {$w_2$};
   \node [m3,anchor=west,fill=blue!30!white] (small24) at ([yshift=-0.0\base]small23.east){};
   \node [font=\small,anchor=south] () at ([yshift=-0.05\base]small24.south) {$w_3$};

   \node [m3,anchor=east,fill=orange!30!white] (small3) at ([xshift=-1.3\base,yshift=0.5\base]small21.west){};
   \node [font=\small,anchor=south] () at ([yshift=-0.05\base]small3.south) {$r_0$};
   \node [font=\small,anchor=south] () at ([yshift=0.1\base]small3.north) {$4$};
   \node [font=\small,anchor=east] () at ([xshift=-0.1\base]small3.west) {$1$};

   \draw [-latex', very thick] ([xshift=0.3\base,yshift=0.0\base]matrix2.east) to ([xshift=1.0\base,yshift=0.0\base]matrix2.east);
   \draw [-latex', very thick] ([xshift=0.0\base,yshift=-0.2\base]matrix4.south) to ([xshift=0.0\base,yshift=-0.7\base]matrix4.south);
   \draw [-latex', very thick] ([xshift=1.0\base,yshift=0.0\base]small3.east) to ([xshift=0.3\base,yshift=0.0\base]small3.east);
    
\end{scope}
\end{tikzpicture}
\caption{An example of processing multiple integers in a single SIMD instruction.}
\label{fig:gemm}
\end{figure}

\subsection{GEMM Optimization}
% \fig{fig:time_piechart} shows the proportion of running time for each operation in the decoding process.
% The corresponding data is counted in 32-bit floating point format on the ONNX Runtime.
According to statistics on the ONNX Runtime platform, general matrix multiplication (GEMM) accounts for 80.44\% of the overall decoding time, demonstrating that optimizing GEMM is the key to decoding speed up.
We optimize GEMM from three aspects: 
(1) Replacing 32-bit floating points with 8-bit integers in GEMM for model quantization. 
(2) The Arm instruction set we use allows multiple integers to be processed in parallel in a single instruction, which takes full advantage of the processor throughput.
(3) To improve the cache hit and the register usage, we adjust the layout of the tensor in memory to ensure that the instruction reads data from continuous space. 
Specifically, we convert each $4\times4$ block in the original layout into a contiguous vector of size $16$.
An example can be seen in \fig{fig:gemm}.

\subsection{Memory Optimization}
As shown in \fig{fig:time_piechart} in the appendix \ref{sec:operations}, except for GEMM, other operations account for only 19.56\% of the decoding time but will be frequently performed, resulting in a large amount of temporary memory.
To improve memory efficiency, we take two strategies: 
(1) To avoid frequent memory-mapped I/O and footprint, our engine integrates all adjacent fine-grained operations between two GEMM operations into one fused operation.
(2) To save temporary memory, different operations are allowed to share the same space, provided that these operations do not interfere with each other at the same time.
Through memory sharing, only two 8-bit memory buffers, and one 32-bit buffer need to be pre-allocated in the Transformer encoder to hold intermediate results.

\section{Experiments}

\begin{table}[!t]
  \centering
  \small
  \renewcommand\arraystretch{1.1}
  \setlength{\tabcolsep}{0.2mm}{
    \begin{tabular}{l|c|c|c|c}
      \hline
      \multicolumn{1}{c|}{\multirow{2}{*}{System}} &
      \multicolumn{2}{c|}{\multirow{1}{*}{Params(M)}} &
      \multicolumn{1}{c|}{\multirow{2}{*}{FLOPs(G)}} &
      \multicolumn{1}{c}{\multirow{2}{*}{BLEU}} \\
      \cline{2-3}
      & w/ & w/o & \\
      \hline
      \multirow{1}{*}{Transformer-base} & 65 & 44 & 1.9 & 27.40 \\
      \hline
      \multirow{1}{*}{DeLighT} & 37 & 31.4 & - & 27.60 \\
      \multirow{1}{*}{Universal Transformer} & N/A & 7.4 & 1.9 & 26.20 \\
      \multirow{1}{*}{Lite Transformer (small)} & N/A & 2.9 & 0.2 & 22.50 \\
      \multirow{1}{*}{Lite Transformer (medium)} & N/A & 11.7 & 0.7 & 25.60 \\
      \multirow{1}{*}{Lite Transformer (big)} & N/A & 17.3 & 1.0 & 26.50 \\
      \multirow{1}{*}{EdgeFormer w/o LA} & N/A & 8.6 & 1.8 & 26.50 \\
      \multirow{1}{*}{EdgeFormer (Adapter-LA)} & N/A & 9.4 & 1.8 & 26.90 \\
      \multirow{1}{*}{EdgeFormer (Prefix-LA)} & N/A & 8.6 & 1.9 & 26.80 \\
      \hline
      \multirow{1}{*}{MobileNMT-10MB} & 10 & 7.9 & 0.3 & 25.08 \\
      \multirow{1}{*}{MobileNMT-20MB} & 20 & 17.7 & 0.6 & 27.09 \\
      \hline
    \end{tabular}
  \caption{The comparison of MobileNMT with other parameter-efficient Transformers, including DeLighT \cite{DBLP:conf/iclr/MehtaGIZH21}, Universal Transformer \cite{DBLP:conf/iclr/DehghaniGVUK19}, Lite Transformer \cite{DBLP:conf/iclr/WuLLLH20} and EdgeFormer \cite{DBLP:conf/emnlp/GeCW22}
  (Parameters w/ or w/o embedding layer are both provided. 
  FLOPs is estimated on a sample with src/tgt length of 30.).}
  \label{tab:comparison}
  }
\end{table}

\subsection{Setups}

We evaluate our methods on two WMT benchmarks.
For the WMT14 En-De task (4.5M pairs), 
we choose \emph{newstest-2013} as the validation set and \emph{newstest-2014} as the test set. 
For the WMT14 En-Fr task (35M pairs), we validate the system on the combination of newstest-2012 and newstest-2013, and test it on newstest-2014.
Details of the architecture were introduced in Section \ref{sec:architecture}, and training hyperparameters were introduced in Section \ref{sec:training}.
For model compression ratio and decoding speed up, we choose Transformer-big as the benchmark (1.0$\times$).
Other details of experimental setups are introduced in Appendix \ref{sec:setups}.

\subsection{Results}
\label{subsec:results}
\tab{tab:main_result} shows the results of different systems on WMT14 En-De and En-Fr.
\tab{tab:comparison} shows the comparison of MobileNMT with other parameter-efficient methods based on Transformer.
MobileNMT-10MB and MobileNMT-20MB are two models we have built with different sizes, which are introduced in \tab{tab:detailed_mobilenmt}.

On WMT14 En-De, our MobileNMT-10MB requires only 4.6\% of the parameters to maintain 88.4\% performance of Transformer-big, while it achieves 87.2$\times$ compression ratio and 47.0$\times$ speed up.
Our MobileNMT-20MB can maintain 95.5\% performance of Transformer-big with only 9.2\% parameters, while it achieves 43.6$\times$ compression ratio and 27.7$\times$ speed up.
Experiments on En-Fr show similar results.
In addition, thanks to the memory optimization strategies adopted in our engine, MobileNMT requires significantly less running memory than other models (0.5\%$\sim$0.9\% of Transformer-big).
All these experiments demonstrate that MobileNMT is efficient in terms of parameters, computation, and memory, and can be easily deployed on mobile devices. 

% \section{Related Work}

\section{Conclusion}

% In this paper, we propose MobileNMT, a hardware-friendly machine translation system.
% It is efficient in terms of parameters, computation, and memory, and can be easily deployed on mobile devices. 
We propose MobileNMT, a Transformer-based machine translation system that can translate in 15MB and 30ms.
It uses existing resources efficiently and can be easily deployed in real-world scenarios.
We develop a mobile inference engine with GEMM and memory optimization, hoping that it can bridge the gap between theoretical research and real-world applications on efficient machine translation.

\section*{Acknowledgments}

% This work was supported in part by the National Science Foundation of China (Nos. 61876035 and 61732005), the China HTRD Center Project (No. 2020AAA0107904), Yunnan Provincial Major Science and Technology Special Plan Projects (Nos. 202002AD080001 and 202103AA080015), National Frontiers Science Center for Industrial Intelligence and Systems Optimization (Northeastern University, China. No. B16009) and the Fundamental Research Funds for the Central Universities.
% The authors would like to thank anonymous reviewers for their comments.
This work was supported in part by the National Science Foundation of China (No. 62276056), the National Key R\&D Program of China, the China HTRD Center Project (No. 2020AAA0107904), the Natural Science Foundation of Liaoning Province of China (2022-KF-16-01), the Yunnan Provincial Major Science and Technology Special Plan Projects (No. 202103AA080015), the Fundamental Research Funds for the Central Universities (Nos. N2216016, N2216001, and N2216002), and the Program of Introducing Talents of Discipline to Universities, Plan 111 (No. B16009).

\section*{Limitations}

\noindent \textbf{Multilingual Translation.}
Here we mainly discuss the design principles of efficient architectures for bilingual machine translation.
Compared with bilingual translation, multilingual translation tasks require significantly more parameters and computations to perform well, and different model scales may lead to different design considerations. 
We will leave this for future exploration.

\noindent \textbf{Knowledge Distillation.}
As a small model that requires only 10MB/20MB of storage, MobileNMT will inevitably suffer from performance loss compared to other Transformer-based models.
To reduce performance loss, here we adopt knowledge distillation and choose the Transformer-base model as the teacher.
From a training efficiency perspective, although the teacher model can help MobileNMT improve performance, it also introduces additional training costs.

\noindent \textbf{Compatibility.}
Here our inference engine only provides implementation for the ARM CPU.
We will make it available for other AI accelerator (such as NPU) on mobile devices in the future.
  
% \section*{Acknowledgments}

\bibliography{anthology,custom}
\bibliographystyle{acl_natbib}

\clearpage

\appendix

\section{Transformer Architecture}
\label{sec:transformer}

We chose Transformer for study because it is one of the most successful neural models for machine translation.
It consists of a $N$-layer encoder and a $M$-layer decoder, where $N$=$M$=6 in the original Transformer-base and Transformer-big. 
Each encoder layer consists of two sublayers, including the self-attention and feed-forward network (FFN).
Each decoder layer has an additional cross-attention sublayer to bridge the encoder and decoder. 

The self-attention takes the output $X$ of the previous sublayer as its input. 
The cross-attention is similar to the self-attention, except that it takes the encoder output as an additional input.
Both types of attention first compute the attention distribution $A_x$ and then average $X$ by $A_x$. 
We denote the transformation matrices of $Q,K,V$ as $W_{q},W_{k},W_{v}$, the subsequent transformation matrices as $W_{o}$, and the attention as $Y_a=\mathrm{Attn}(X)$, then:
\begin{align}
    A_x&=\mathrm{SoftMax}(\frac{XW_{q}W_{k}^TX^T}{\sqrt{d}})\label{eqn:self-weight}\\
    Y_a&=A_xXW_{v}W_{o}\label{eqn:self-sum}
\end{align}

The FFN applies non-linear transformation to its input $X$. 
We denote the FFN as $Y_f=\mathrm{FFN}(X)$:
\begin{equation}
    Y_f=\mathrm{ReLU}(XW_1+b_1)W_2+b_2\label{eqn:ffn}
\end{equation}
where $W_1$ and $b_1$ denote the weight and bias of the first linear transformation, $W_2$ and $b_2$ are parameters of the second transformation.

Here we preprocess each sublayer input by the layer normalization \cite{DBLP:journals/corr/BaKH16}.
All sublayers are coupled with the residual connection \cite{DBLP:conf/cvpr/HeZRS16}.

% For more details, we refer the reader to \citet{DBLP:conf/nips/VaswaniSPUJGKP17}.

\begin{figure*}[t]
  \centering
  \small
  \begin{tikzpicture}
    \begin{axis}[
      width=12.0cm, height=9.0cm, 
      xbar, 
      grid style=dashed,
      ymajorgrids=true,
      xmajorgrids=true,
      ylabel near ticks,
      xlabel near ticks,
      xlabel={Percentage [\%]},
      xmin=0.0,xmax=0.07,
      bar width=8pt,
      symbolic y coords={IDENTITY,SLICE,CUMSUM,EXPAND,WHERE,MULTI\_HEAD,HOST\_CONVERTER,SHAPE,SPLIT,UNSQUEEZE,CONCAT,CONSTANT\_OF\_SHAPE,MULTI\_HEAD\_TRANS,BINARY\_MUL,REDUCE,BINARY\_UNKNOWN,GATHER,CONVERTER,BINARY\_ADD,LAYER\_NORM,ARGMAX,DECODER\_ATTENTION,DECODER\_SELF\_ATTENTION,ENCODER\_SELF\_ATTENTION,GEMM},
      ytick=data,
      nodes near coords,
      nodes near coords style={font=\small}]
      \addplot[draw=lyyblue,fill=lyyblue!90] coordinates {
      (2.09E-06,IDENTITY)(2.55E-06,SLICE)(4.79E-06,CUMSUM)(1.22E-05,EXPAND)(2.45E-05,WHERE)(5.52E-05,MULTI\_HEAD)(6.28E-05,HOST\_CONVERTER)(7.10E-05,SHAPE)(9.49E-05,SPLIT)(0.000112952,UNSQUEEZE)(0.000151531,CONCAT)(0.000191037,CONSTANT\_OF\_SHAPE) (0.000200237,MULTI\_HEAD\_TRANS)(0.000200856,BINARY\_MUL) (0.000356562,REDUCE) (0.000412845,BINARY\_UNKNOWN) (0.000463097,GATHER)(0.00298655,CONVERTER) (0.00488092,BINARY\_ADD) (0.00535685,LAYER\_NORM) (0.0188322,ARGMAX)(0.0196205,DECODER\_ATTENTION)(0.0376169,DECODER\_SELF\_ATTENTION) (0.0589383,ENCODER\_SELF\_ATTENTION)};
      % (0.804412,GEMM) 
    \end{axis}
  \end{tikzpicture}
  \caption{Proportions of different operations (except GEMM) on the Transformer-base model.}
  \label{fig:time_piechart}
\end{figure*}
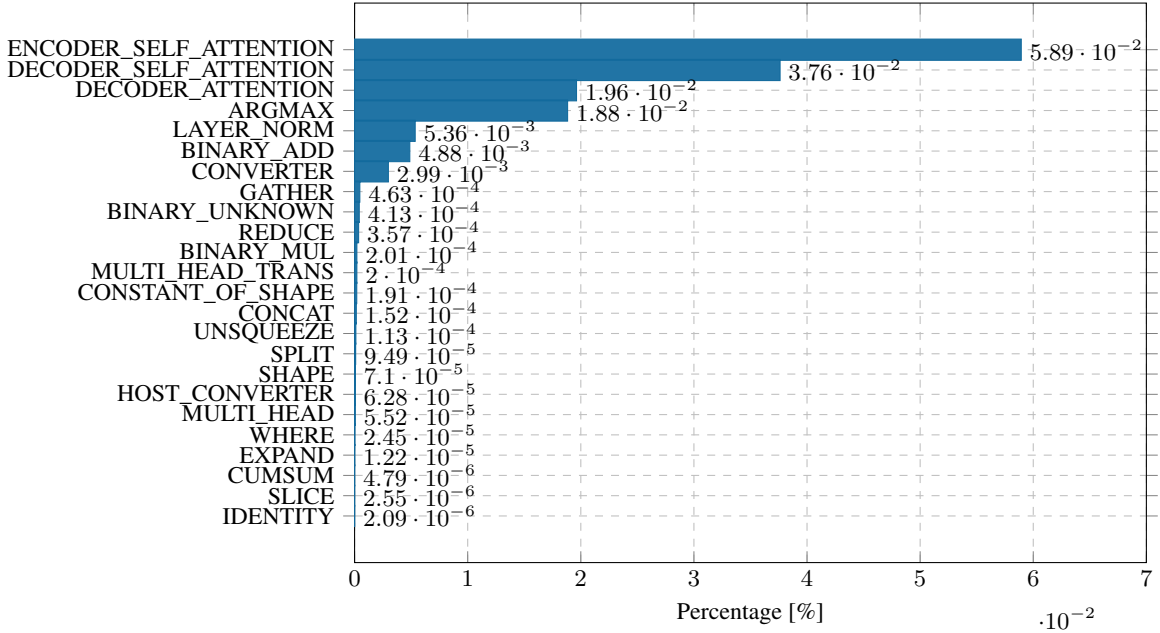

% \section{Details of Base, Small and Tiny}
% \label{sec:setting}
% % To compare the two embedding compression methods mentioned in Section \ref{sec:architecture}, 
% In this paper, we select three baseline models of different sizes. 
% Detailed model settings are shown in \tab{tab:detailed}.

\section{PTQ and QAT}
\label{sec:ptq_qat}

As an appealing solution to model compression, quantization enables the model to use lower-bit values (such as 8-bit integer) to compute faster and consume less storage space \cite{DBLP:conf/nips/HubaraCSEB16,DBLP:conf/iclr/MicikeviciusNAD18,DBLP:conf/naacl/QuinnB18,DBLP:conf/cvpr/JacobKCZTHAK18}.

Post-Training Quantization (PTQ) can be seen as the basis for Quantization Aware Training (QAT), it adds quantization nodes to a well-trained floating-point model.
To quantize a floating-point tensor $r$ to a tensor with $n$ bits, a scale $s$ is introduced to map these two types of values \cite{DBLP:journals/corr/abs-2001-00926}:
\begin{align}
    s&=\frac{\max(r)-\min(r)}{2^n-1}
    \label{eqn:scale}
\end{align}

To get a faster computation speed, both weights and activations will be quantized to $n$-bit.
Suppose $r_m=\min(r)$, the quantization function is:
\begin{align}
    Q(r)=\left\lfloor(r-r_m)/s\right\rceil\times s+r_m
    \label{eqn:quant}
\end{align}
where $\left\lfloor \cdot \right\rceil$ represents rounding to the nearest integer.

However, in PTQ, applying quantization directly to the floating-point network will result in significant performance losses.
Based on PTQ, QAT simulates the behavior of $n$-bit computation by minimizing quantization errors during training, which helps the model achieve higher accuracy.
In addition to the learnable weights of the model itself, $s$ is also learnable.

\section{Operations except GEMM}
\label{sec:operations}

Since general matrix multiplication (GEMM) accounts for 80.44\% of the overall decoding time, we have concluded that optimizing GEMM is the key to decoding speed up in Section \ref{sec:engine}.
As for operations except GEMM, \fig{fig:time_piechart} shows the proportion of running time in the decoding process.
The corresponding data is measured in 32-bit floating point format on the ONNX Runtime.

% \begin{figure*}[t!]
% \centering
% \small
% \includegraphics[width=0.95\linewidth]{engine.png}
% \caption{Implementation details of the engine.}
% \label{fig:engine}
% \end{figure*}

\begin{table}[!t]
  \centering
  \small
  \renewcommand\arraystretch{1.1}
  \setlength{\tabcolsep}{1.4mm}{
    \begin{tabular}{l|c|c|c}
      \hline
      \multicolumn{1}{c|}{\multirow{1}{*}{System}} &
      \multicolumn{1}{c|}{\multirow{1}{*}{Params (M)}} &
      \multicolumn{1}{c|}{\multirow{1}{*}{Size (MB)}} &
      \multicolumn{1}{c}{\multirow{1}{*}{BLEU}} \\
      % \cline{4-6}
      % & (M) & (MB) & \\
      \hline
      \multirow{1}{*}{Transformer-base} & 65 & 260 & 27.40 \\
      \hline
      \rowcolor{lyyellow}
      \multirow{1}{*}{+ Reducing Vocab} & 48 & 192 & 26.20 \\
      \rowcolor{lyyellow}
      \multirow{1}{*}{+ Reducing Width} & 10 & 40 & 22.01 \\
      \rowcolor{lyyellow}
      \multirow{1}{*}{+ Other Dimensions} & 10 & 40 & 22.54 \\
      \rowcolor{lyblue}
      \multirow{1}{*}{+ Distillation} & 10 & 40 & 23.77 \\
      \rowcolor{lyblue}
      \multirow{1}{*}{+ Quantization} & 10 & 10 & 23.76 \\
      \rowcolor{lyblue}
      \multirow{1}{*}{+ Hyperparameters} & 10 & 10 & 25.48 \\
      % \rowcolor{lypink}
      \multirow{1}{*}{+ Greedy Search} & 10 & 10 & 25.08 \\
      \hline
    \end{tabular}
  \caption{Ablation study on MobileNMT-10MB. The colors refer to \colorbox{lyyellow}{Model Architecture} in Section \ref{sec:architecture}, \colorbox{lyblue}{Training Strategies} in Section \ref{sec:training} and Greedy Search.}
  \label{tab:ablation_study}
  }
\end{table}

\section{Setups}
\label{sec:setups}

All sentences were segmented into sequences of sub-word units \cite{DBLP:conf/acl/SennrichHB16a}.
In the implementation, we adopt the normalization before layers  \cite{DBLP:conf/iclr/BaevskiA19,DBLP:conf/icml/XiongYHZZXZLWL20,DBLP:journals/corr/abs-1910-05895}.
Most previous work only shared source and target vocabularies on the En-De task.
In our MobileNMT, both En-De and En-Fr adopt shared vocabularies for efficiency reasons, which leads to a larger compression gain at the expense of performance. 
We test on the model ensemble by averaging the last 5 checkpoints and report  SacreBLEU scores \cite{DBLP:conf/wmt/Post18}.

For the experiments of MobileNMT in \tab{tab:main_result}, we use the greedy search algorithm in our engine.
Compared with beam search, greedy search can lead to more efficient decoding.
For the experiments of TFLite in \tab{tab:main_result}, since TFLite will expand all loop subgraphs, it is hard to support the entire decoding process (30 steps) of the Transformer-big/base model with limited memory (6GB in Google Pixel 4). 
For the memory of these two models, we only record the running memory of 1 step.
For the corresponding latencies, we estimate the 30-step latency according to the 1-step and 5-step latencies.
It is worth noting that except for the memory and latency on Transformer-big/base, all other data statistics are measured in real-world.

\section{Analysis}

\begin{table}[!t]
  \centering
  \small
  \renewcommand\arraystretch{1.1}
  \setlength{\tabcolsep}{1.7mm}{
    \begin{tabular}{l|c|c|c|c}
      \hline
      \multicolumn{1}{c|}{\multirow{2}{*}{System}} &
      \multicolumn{1}{c|}{\multirow{1}{*}{Params}} &
      \multicolumn{1}{c|}{\multirow{1}{*}{Bits}} &
      \multicolumn{1}{c|}{\multirow{1}{*}{Size}} &
      \multicolumn{1}{c}{\multirow{2}{*}{BLEU}} \\
      % \cline{4-6}
      & (M) & (W-E-A) & (MB)  \\
      \hline
      \multirow{1}{*}{Transformer-base} & 65 & 32-32-32 & 260 & 27.40 \\
      \hline
      \multirow{5}{*}{MobileNMT-10MB} & 10 & 32-32-32 & 40 & 25.79 \\
      & 10 & 8-8-8 & 10 & 25.08 \\
       & 10 & 4-8-8 & 5 & 25.43 \\
      & 10 & 3-8-8 & 3.75 & 24.09 \\
      & 10 & 2-8-8 & 2.5 & 21.25 \\
      \hline
      \multirow{5}{*}{MobileNMT-20MB} & 20 & 32-32-32 & 80 & 27.30 \\
      & 20 & 8-8-8 & 20 & 27.09 \\
      & 20 & 4-8-8 & 10 & 26.96 \\
      & 20 & 3-8-8 & 7.5 & 26.23 \\
      & 20 & 2-8-8 & 5 & 24.33 \\
      \hline
    \end{tabular}
  \caption{Results of quantizing weights to lower bits.}
  \label{tab:quantization}
  }
\end{table}

\subsection{Ablation Study}

\tab{tab:ablation_study} summarizes how each part of Section \ref{sec:architecture} and Section \ref{sec:training} affects the overall performance.
Each row in \tab{tab:ablation_study} represents the result of applying the current part to the system obtained in the previous row.

To reduce the model parameters from 65M to 10M, the model performance decreased from 27.40 to 22.54, which illustrates the importance of network parameters on model capacity. 
We observe that both knowledge distillation and tuning hyperparameters can bring significant performance improvements (from 22.54 to 25.48), which effectively compensate for the performance loss caused by parameter reduction.

\subsection{Quantization Study}
\tab{tab:quantization} studies how performance changes when quantizing the model to lower bits (i.e., 4-bit, 3-bit, and 2-bit).
As introduced in Section \ref{subsec:quantization}, for 8-bit quantization, we use the element-wise quantization method \cite{DBLP:conf/cvpr/Lee0H21}.
For lower-bit quantization, we use the row-wise quantization for accuracy preservation \cite{DBLP:conf/cvpr/FaraoneFBL18}.

As shown in \tab{tab:quantization}, 8-bit and 4-bit quantization have almost no negative effect on model performance.
When quantizing the model to lower bits, such as 3-bit and 2-bit integers, model performance will drop dramatically.

\end{document}